\documentclass[pdflatex,sn-mathphys]{sn-jnl}
\usepackage[T1]{fontenc}
\usepackage{lmodern}
\usepackage{fix-cm}

\jyear{2023}%

\theoremstyle{thmstyleone}%
\theoremstyle{thmstyletwo}%

\theoremstyle{thmstylethree}%
\makeatletter
\renewcommand\normalsize{%
  \@setfontsize\normalsize{9pt}{11pt}
}
\makeatother

\usepackage{mathtools}        
\usepackage{amssymb,amsfonts}
\usepackage{amsthm}
\usepackage{mathrsfs}

\usepackage{graphicx}
\usepackage{xcolor}

\usepackage{booktabs,multirow}        
\usepackage{makecell}                 
\usepackage{array,tabularx}           
\usepackage{threeparttable}           
\usepackage{rotating}                 
\usepackage{eqparbox}                 
\usepackage{soul}                     
\usepackage{siunitx}                  
\usepackage{xurl}
\usepackage{pdflscape}
\usepackage{rotating}

\usepackage{algorithm}       
\usepackage{algpseudocode}
\algnewcommand\algorithmicbreak{\textbf{break}}
\algnewcommand\Break{\State \algorithmicbreak}
\algrenewcommand\algorithmicrequire{\textbf{Input:}}
\algrenewcommand\algorithmicensure{\textbf{Output:}}
\makeatletter
\newcommand{\NoIndentState}{\State\hspace*{-\ALG@thistlm}}
\makeatother
\usepackage{float}              
\usepackage[section]{placeins}  
\usepackage[font=small,labelfont=bf]{caption}

\usepackage{listings}
\usepackage[title]{appendix}
\usepackage{textcomp}
\usepackage{manyfoot}
\usepackage{bbm}
\usepackage{ragged2e}                 
\usepackage{longtable}

\usepackage{graphicx}
\hypersetup{colorlinks=true, breaklinks=true}

\usepackage{fontawesome5}

\hypersetup{colorlinks=true, urlcolor=blue}
\usepackage{lineno}

\begin{document}

\title[OmniCellTOSG]{OmniCellTOSG: The First Cell Text-Omic Signaling Graphs Dataset for Graph Language Foundation Modeling}


\author[1]{\fnm{Heming} \sur{Zhang}}
\equalcont{These authors contributed equally to this work.}

\author[1]{\fnm{Tim} \sur{Xu}}
\equalcont{These authors contributed equally to this work.}

\author[1]{\fnm{Dekang} \sur{Cao}}

\author[1]{\fnm{Shunning} \sur{Liang}}

\author[1]{\fnm{Guntaas} \sur{Shergill}}

\author[1]{\fnm{Nicholas} \sur{Hadas}}

\author[1]{\fnm{Lars} \sur{Schimmelpfennig}}

\author[1]{\fnm{Levi} \sur{Kaster}}

\author[1,2]{\fnm{Di} \sur{Huang}}

\author[9]{\fnm{Guangfu} \sur{Li}}

\author[5,6]{\fnm{S. Peter} \sur{Goedegebuure}}

\author[6,7]{\fnm{David} \sur{DeNardo}}

\author[6,7]{\fnm{Li} \sur{Ding}}

\author[5,6]{\fnm{Ryan C.} \sur{Fields}}

\author[1]{\fnm{J Philip} \sur{Miller}}

\author[8]{\fnm{Pirooz} \sur{Eghtesady}}

\author[3]{\fnm{Carlos} \sur{Cruchaga}}

\author[4]{\fnm{William} \sur{Buchser}}

\author[8]{\fnm{Jonathan} \sur{Cooper}}

\author[8]{\fnm{Marco} \sur{Sardiello}}

\author[8]{\fnm{Patricia} \sur{Dickson}}

\author[2]{\fnm{Yixin} \sur{Chen}}

\author[4]{\fnm{Michael} \sur{Province}}

\author[1]{\fnm{Philip} \sur{Payne}}

\author*[1,2,8]{\fnm{Fuhai} \sur{Li}}\email{fuhai.li@wustl.edu}

\affil[1]{\orgdiv{Institute for Informatics, Data Science and Biostatistics}, \orgname{Washington University in St. Louis}, \orgaddress{\street{4444 Forest Park Ave.}, \city{Saint Louis}, \postcode{63108}, \state{MO}, \country{USA}}}

\affil[2]{\orgdiv{Department of Computer Science}, \orgname{Washington University in St. Louis}, \orgaddress{\street{1 Brookings Dr.}, \city{Saint Louis}, \postcode{63130}, \state{MO}, \country{USA}}}

\affil[3]{\orgdiv{NeuroGenomics and Informatics Center}, \orgname{Washington University in St. Louis}, \orgaddress{\street{4444 Forest Park Ave.}, \city{Saint Louis}, \postcode{63108}, \state{MO}, \country{USA}}}

\affil[4]{\orgdiv{Department of Genetics}, \orgname{Washington University in St. Louis}, \orgaddress{\street{4515 McKinley Ave.}, \city{Saint Louis}, \postcode{63110}, \state{MO}, \country{USA}}}

\affil[5]{\orgdiv{Department of Surgery}, \orgname{Washington University in St. Louis}, \orgaddress{\street{4444 Forest Park Ave.}, \city{Saint Louis}, \postcode{63110}, \state{MO}, \country{USA}}}

\affil[6]{\orgdiv{Siteman Cancer Center}, \orgname{Washington University in St. Louis}, \orgaddress{\street{4444 Forest Park Ave.}, \city{Saint Louis}, \postcode{63110}, \state{MO}, \country{USA}}}

\affil[7]{\orgdiv{Department of Medicine}, \orgname{Washington University in St. Louis}, \orgaddress{\street{4444 Forest Park Ave.}, \city{Saint Louis}, \postcode{63110}, \state{MO}, \country{USA}}}

\affil[8]{\orgdiv{Department of Pediatrics}, \orgname{Washington University in St. Louis}, \orgaddress{\street{4444 Forest Park Ave.}, \city{Saint Louis}, \postcode{63110}, \state{MO}, \country{USA}}}

\affil[9]{\orgdiv{Department of Surgery}, \orgname{University of Connecticut}, \orgaddress{\street{263 Farmington Ave.}, \city{Farmington}, \postcode{06032}, \state{CT}, \country{USA}}}


\abstract{
With the rapid growth of large-scale single-cell omic datasets, omic foundation models (FMs) have emerged as powerful tools for advancing research in life sciences and precision medicine. However, most existing omic FMs rely primarily on numerical transcriptomic data by sorting genes as sequences, while lacking explicit integration of biomedical prior knowledge and signaling interactions that are critical for scientific discovery. Here, we introduce the Text-Omic Signaling Graph (TOSG), a novel data structure that unifies human-interpretable biomedical textual knowledge, quantitative omic data, and signaling network information. Using this framework, we construct OmniCellTOSG, a large-scale resource comprising approximately half million meta-cell TOSGs derived from around 80 million single-cell and single-nucleus RNA-seq profiles across organs and diseases. We further develop CellTOSG-FM, a multimodal graph language FM, to jointly analyze textual, omic and signaling network context. Across diverse downstream tasks, CellTOSG-FM outperforms existing omic FMs, and provides interpretable insights into disease-associated targets and signaling pathways. 

}



\maketitle

\section{Main}\label{main}
The human organism comprises $\sim$37.2 trillion cells that arise from a single zygote and share a common genome, yet acquire specialized identities through context-dependent signaling. Such signaling is orchestrated by transcriptional programs, protein abundance and modification, and protein--protein interactions, and is further conditioned by age, sex, diet, environmental exposures, and disease state. Despite decades of discovery, major gaps persist: system-level, cell-resolved inventories of signaling entities and edges; quantitative models of network rewiring across lifespan and pathology; principled detection of disease-relevant subpopulations and their intercellular crosstalk; and actionable strategies to perturb these networks to prevent or reverse disease. Single-cell and single-nucleus RNA sequencing (sc/snRNA-seq) now provide transcriptome-wide measurements at cellular resolution, enabling delineation of cell types/subtypes in healthy and diseased tissues and the study of signaling interactions within niches or microenvironments. Large-scale efforts, such as the CZ CELLxGENE~\cite{megill2021cellxgene,czi2025cz}, the Human Cell Atlas~\cite{rood2024human}, the Brain Cell Atlas~\cite{chen2024brain}, and numerous disease-focused studies~\cite{miller2023sea, mathys2023single}, have generated hundreds of millions of profiles that support systematic interrogation of signaling. These resources make it feasible to ask not only which genes are active, but how groups of genes/proteins with distinct abundance levels coordinate to realize specific biological functions across diverse cellular contexts.

Currently, foundation models trained via self-supervised objectives have transformed representation learning. Most foundation single-cell models operate on expression vectors and typically do not incorporate explicit pathway structure, including SCimilarity~\cite{heimberg2024cell}, GeneFormer~\cite{theodoris2023transfer}, scGPT~\cite{cui2024scgpt}, scFoundation~\cite{hao2024large}, and scCello~\cite{yuan2024cell}. Despite recent progress, these approaches generally omit explicit modeling of signaling graphs, limiting inference of dysfunctional pathways and decoding of graph-structured signaling patterns across conditions. Recent progress on training graph foundation models has explored masked reconstruction objectives within masked graph modeling, with node-masking methods such as GraphMAE~\cite{hou2022graphmae} as representative examples. Systematic analysis suggests that masking edges rather than nodes yields stronger performance on structure-sensitive tasks, including link prediction and topology recovery, and better captures relational patterns~\cite{li2023s}. This is particularly salient for cellular signaling, where functional meaning arises from interaction topology rather than isolated node attributes. Interpreting these structure-dependent mechanisms typically requires both biomedical prior knowledge and topological information, yet existing models remain limited in both aspects: LLMs often struggle with domain-specific biomedical reasoning and may produce hallucinated or unreliable outputs~\cite{bender2021dangers}, whereas GNNs can be limited in modeling complex higher-order relational structures~\cite{dong2023rethinking,abboud2020surprising}. In addition to these limitations, purely numeric omic-based foundation models typically treat molecular measurements as isolated features and rarely incorporate human-interpretable textual biomedical priors and signaling-network context, which limits mechanistic interpretability and hypothesis-driven discovery. Prior work suggests that integrating biologically grounded knowledge graphs with quantitative omic features can improve predictive accuracy~\cite{zhang2024graphseqlm} and strengthen mechanistic reasoning by explicitly capturing cellular interactions~\cite{zhang2025galax}. Collectively, these challenges motivate the development of a unified representation that jointly incorporates textual priors, omic evidence, and signaling topology.

In this study, for the first time, we introduce (i) \textbf{Text--Omic Signaling Graphs (TOSGs)}, a novel data format that unifies textual biological priors (e.g., gene/protein functions, mechanisms) with numerical omic data to support graph-based interpretation of cell signaling; (ii) \textbf{OmniCellTOSG}, a large-scale biomedical AI resource aggregating approximately 80 million sc/snRNA-seq profiles across tissues, cell types, diseases, ages, sexes, and related attributes, providing a comprehensive data foundation to support the development and benchmarking of next-generation AI foundation models for scientific discovery at an expert level; and (iii) \textbf{CellTOSG Foundation Model (CellTOSG-FM)}, a multi-modal graph language foundation model that couples textual biological priors and numerical omic evidence with topological signaling network over TOSGs and cross-modalities encoders to augment graph representation learning and to support downstream tasks, including cell-type annotation, cell attribute classification, and signaling inference with interpretable graph rationales (iv) All the data and code are publicly accessible. OmniCellTOSG dataset is accessible at:
\href{https://huggingface.co/datasets/FuhaiLiAiLab/OmniCellTOSG_Dataset}{\texttt{huggingface.co/datasets/FuhaiLiAiLab/OmniCellTOSG\_Dataset}} and 
CellTOSG-FM code is available at: \href{https://github.com/FuhaiLiAiLab/OmniCellTOSG}{\texttt{github.com/FuhaiLiAiLab/OmniCellTOSG}}

\section{Results}\label{result}
\subsection{OmniCellTOSG ecosystem overview}
\begin{figure}[h]
\vspace{-0.00in}
\centering
\captionsetup{skip=0.1pt} 
\includegraphics[width=0.95\textwidth]{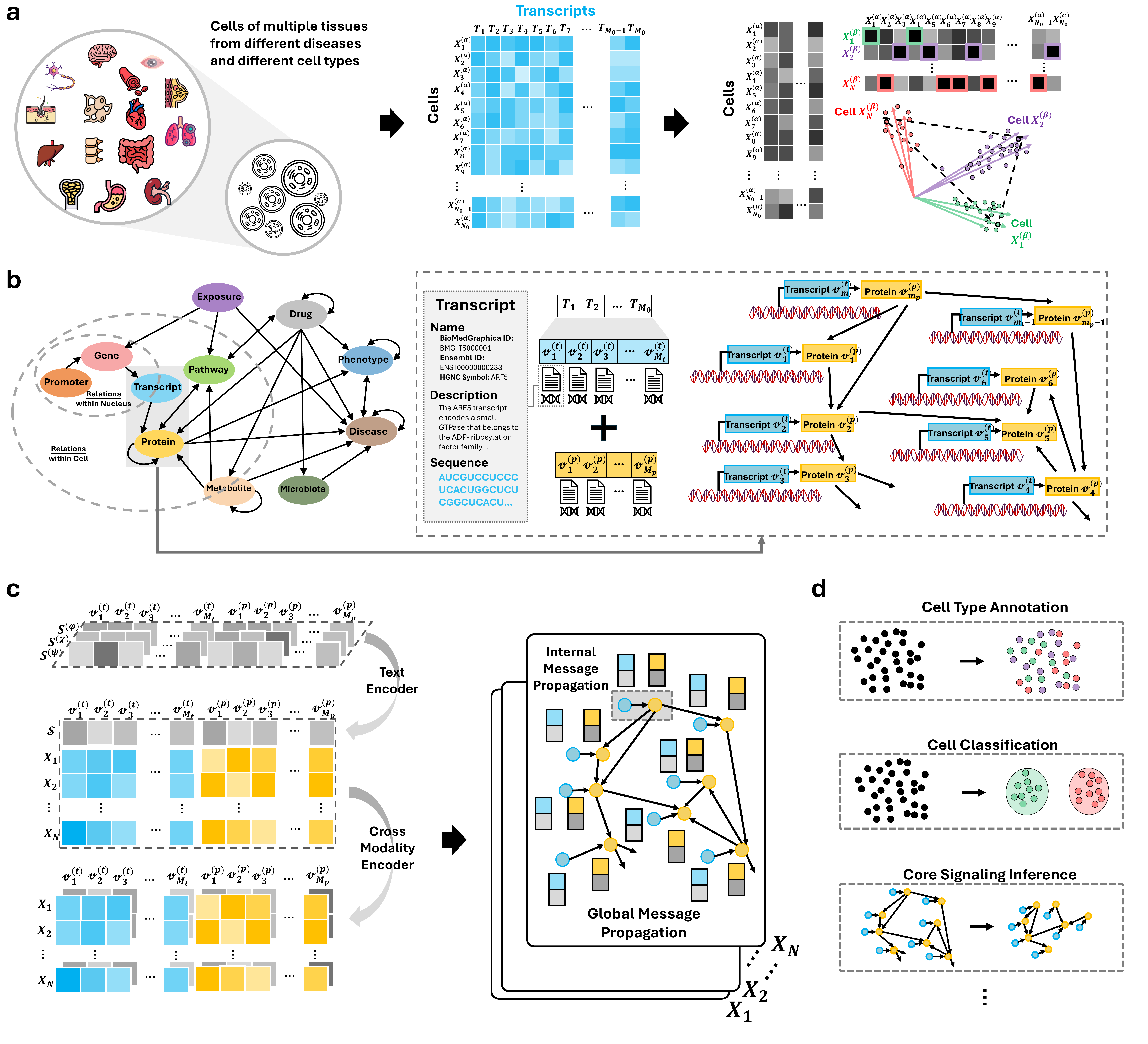}
\caption{\textbf{OmniCellTOSG: data construction and foundation model pipeline.}
\textbf{(a)} Millions of sc/snRNA-seq profiles from diverse tissues, diseases, and cell types (h5ad) are integrated; transcript matrices are extracted from h5ad files and $N_0$ cells are transformed into $N$ meta-cells via archetypal analysis.
\textbf{(b)} Knowledge-grounded graph assembly: transcript and its downstream protein entities are aligned to the BioMedGraphica knowledge base ($M=M_t+M_p$) to form Text–Omic Signaling Graphs (TOSGs) with both matched and virtual entities, capturing nucleus-level and intra-cell relations.
\textbf{(c)} Cross-modal representation learning: a language-modality encoder embeds biological priors (entity names, descriptions, sequences) and is fused with omic features; message passing operates within cells (internal) and across the TOSG (global) to yield unified representations for pretraining and downstream tasks.
\textbf{(d)} Example downstream tasks enabled by the pretrained model: cell-type annotation, disease classification, core-signaling inference, etc.}
\label{fig:architecture}
\vspace{-0.05in}
\end{figure}
We present an integrated ecosystem that couples a large, knowledge-grounded single-cell resource with reproducible tooling and a multi-modal graph language foundation model (see Figure~\ref{fig:architecture}). \textbf{OmniCellTOSG} aggregates approximately $N_0$ ($N_0\simeq80$ million) million single-cell and single-nucleus RNA sequencing profiles into $N$ ($N\simeq0.5$ million) representative meta-cells using the archetypal analysis framework implemented in SEACells~\citep{persad2023seacells}. This aggregation preserves biological diversity across tissues, diseases, age groups, and experimental conditions (Figure~\ref{fig:architecture}a). Based on the resulting meta-cell transcriptomic expression matrices, we next construct \textbf{Text--Omic Signaling Graphs (TOSGs)} by mapping the $M_0$ transcriptomic entities to $M_t$ transcript nodes and introducing $M_p$ corresponding protein nodes, with edges linking transcript entities to their associated proteins. The resulting graph therefore contains a total of $M$ nodes, where $M = M_t + M_p$ (see Section~\ref{data_processing} for details). These graphs integrate quantitative omics measurements with curated biological knowledge from BioMedGraphica on the vertex set $\mathcal{V}$ (see Section~\ref{tosg_generation}). Formally, the overall entity set is defined as
$\mathcal{V} = \{\mathcal{V}^{(t)},\, \mathcal{V}^{(p)}\}$, with $\lvert \mathcal{V} \rvert = \lvert \mathcal{V}^{(t)} \rvert + \lvert \mathcal{V}^{(p)} \rvert = M_t + M_p = M$, thereby linking molecular signals with established biomedical prior knowledge. TOSG supports both matched and virtual entities and records intra-cell and nucleus-level relations, enabling graph-structured signaling beyond expression vectors alone (see Figure~\ref{fig:architecture}b). 
For each entity, numerical omics features are used to form a unified representation. Transcript nodes contain measured transcriptomic expression values, while virtual protein nodes are zero-initialized because no proteomic measurements are available. These features are assembled into a global omics feature matrix $\mathcal{X}$.  In addition to the transcriptomic and virtual proteomic features, we incorporate an auxiliary textual annotation dataset, $\mathcal{S} = \{S^{(\varphi)}, S^{(\chi)}, S^{(\psi)}\}$, which provides complementary semantic information of entity name, description and biosequences (i.e., RNA sequences and protein sequences) for each node (see Section~\ref{tosg_generation} and Figure~\ref{fig:architecture}c for details). Moreover, \texttt{CellTOSG\_Loader} provides a NumPy-ready query–load–balance pipeline that constructs stratified, unbiased cohorts across user-specified facets (cell type, tissue, disease, data source, age/sex, etc.), mitigates class imbalance for pretraining and downstream evaluation.  In addition, \textbf{CellTOSG-FM} integrates textual biological priors and omic features through cross-modal encoders. The graph encoder then performs message propagation within cells and across TOSGs to produce fused representations that enable cell-type annotation, disease classification, signaling-pathway inference, and drug-response prediction, together with interpretable subgraph rationales (see Section~\ref{pretrain} and Figure~\ref{fig:architecture}c for details). Collectively, these components constitute a scalable, mechanism-focused framework that standardizes data ingestion, supports fair and reproducible experimentation, and facilitates knowledge-augmented modeling of cellular signaling at scale. The subsequent sections detail each component.

\subsection{OmniCellTOSG Dataset}
\begin{figure}[h]
\vspace{-0.0in}
\centering
\captionsetup{skip=2.5pt} 
\includegraphics[width=1.0\textwidth]{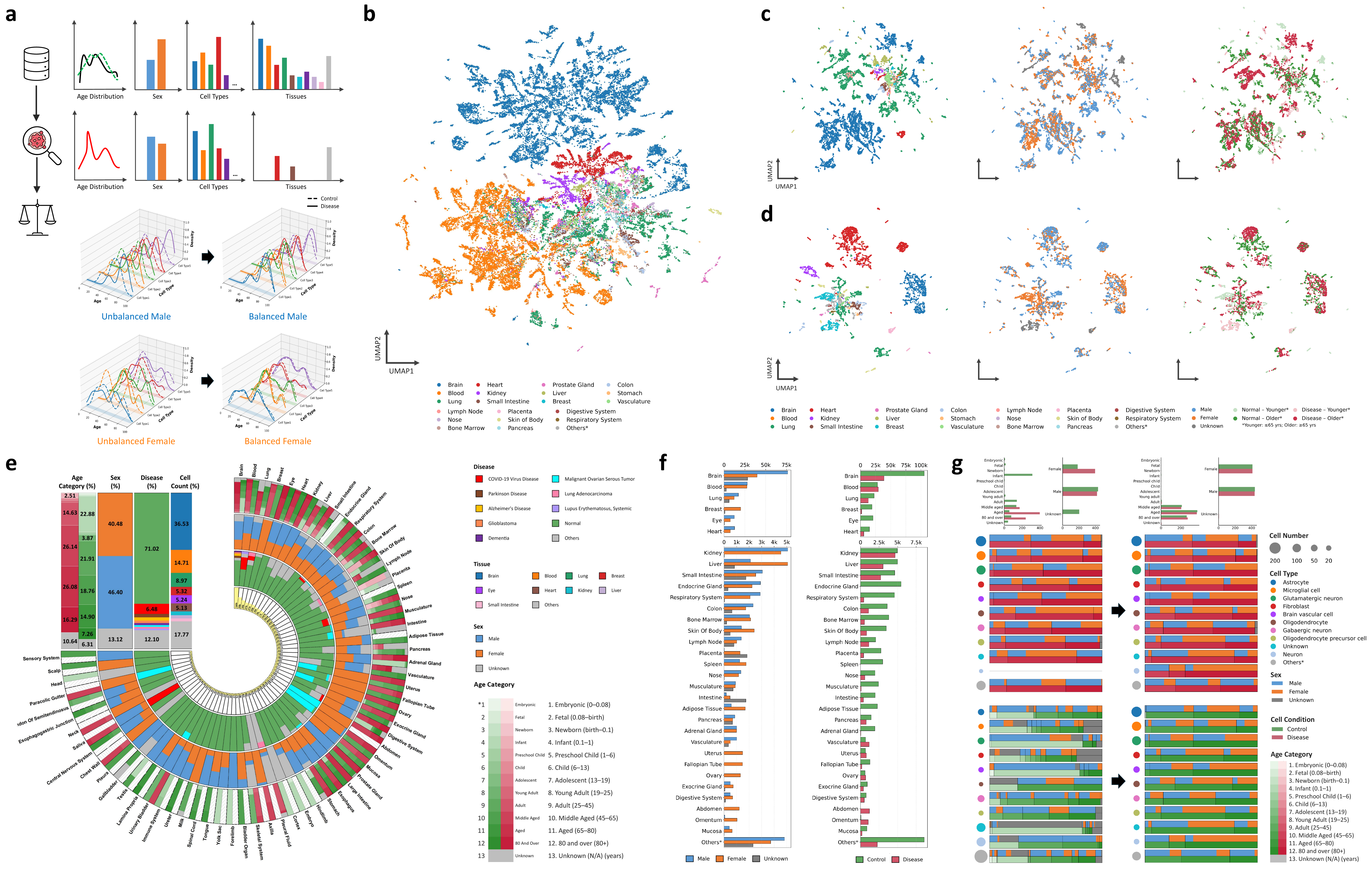}
\caption{\textbf{OmniCellTOSG dataset composition and balancing demo.}
\textbf{(a)} Three-step balancing workflow: first, the raw dataset and its age/sex/cell-type/tissue distributions; second, the retrieved disease-specific distribution used as the target; third, the balancing step and the resulting distributions, such as density overlays for male and female subsets.
\textbf{(b)} UMAP embedding of representative sampled cells colored by types of tissue of origin, highlighting comprehensive inclusion.
\textbf{(c-d)} UMAPs of macrophage cells (upper panel) and fibroblast cells (lower panel) colored by tissue of origin, sex, age groups and disease conditions.
\textbf{(e)} Global composition of the integrated cohort shown as concentric summaries stratified by disease, tissue, sex, age group, and cell counts.
\textbf{(f)} Exact cell counts per tissue, stratified by sex and condition (control vs.\ disease).
\textbf{(g)} Alzheimer’s disease case study: before/after balancing comparisons of the retrieved-and-loaded dataset using stacked bars of cell-type composition, with overlays indicating cell number, sex proportion, condition, and age category proportion.
}
\label{fig:balancing_overview}
\vspace{-0.25in}
\end{figure}

We introduce \textbf{OmniCellTOSG}, a large-scale single-cell resource that integrates transcriptomic profiles from CellxGene~\cite{megill2021cellxgene,czi2025cz}, the Brain Cell Atlas~\cite{chen2024brain}, GEO~\cite{edgar2002gene}, Single Cell Portal~\cite{tarhan2023single}, and the Human Cell Atlas~\cite{regev2017human}, paired with rich textual annotations spanning diverse tissues and disease states. Starting from 79,195,364 cells, we performed rigorous preprocessing—including quality control, normalization, and harmonization of organ/tissue and disease labels. Cells were aggregated into meta-cells using SEACells~\cite{persad2023seacells} and coupled with prior biological knowledge from BioMedGraphica~\cite{zhang2024biomedgraphica} to assemble Text--Omic Signaling Graphs (TOSGs), yielding a curated set of 395,317 meta-cells. Attribute sets were standardized to the Cell Ontology \cite{diehl2016cell} (766 cell types across 65 tissues) and disease annotations were mapped to the BioMedGraphica nomenclature (140 disease states), with remaining fields normalized for retrieval metadata. Following profile harmonization, transcriptomic data were linked to transcript entities and their downstream protein counterparts in BioMedGraphica to construct TOSGs with both matched and virtual entities, capturing nucleus-level and intra-cell relationships. In total, the graph comprises 533,458 entities and 16,637,405 relations (152,585 internal interactions and 16,484,820 protein--protein interactions). A high-level overview of the integrated resource, OmniCellTOSG, is presented in Figure~\ref{fig:balancing_overview}b, and the full data-processing methodology is detailed in Section~\ref{data_processing}.

To ensure reproducibility and a model-ready data format, we release \texttt{CellTOSG\_Loader} (Section~\ref{tosg_loader}), which transforms user-specified parameters into executable queries, loads matched subsets, and performs stratified cohort balancing to mitigate confounding. The loader further applies platform-aware and sc/snRNA-aware batch correction via ComBat--seq~\cite{zhang2020combat,behdenna2023pycombat} to reduce variance across data sources and profiling platforms (Figure~\ref{fig:combat-results}). As an illustrative use case, for an Alzheimer’s disease (AD) versus control comparison, the loader matches the control cohort to the AD distribution over sex, age categories, and cell-type composition. Figure~\ref{fig:balancing_overview}a outlines the balancing workflow, and Figure~\ref{fig:balancing_overview}g shows before/after distributions for AD versus matched normals.

\subsection{CellTOSG-FM Construction and Pretraining}
\begin{figure}[h]
\vspace{-0.0in}
\centering
\captionsetup{skip=2.5pt} 
\includegraphics[width=0.97\textwidth]{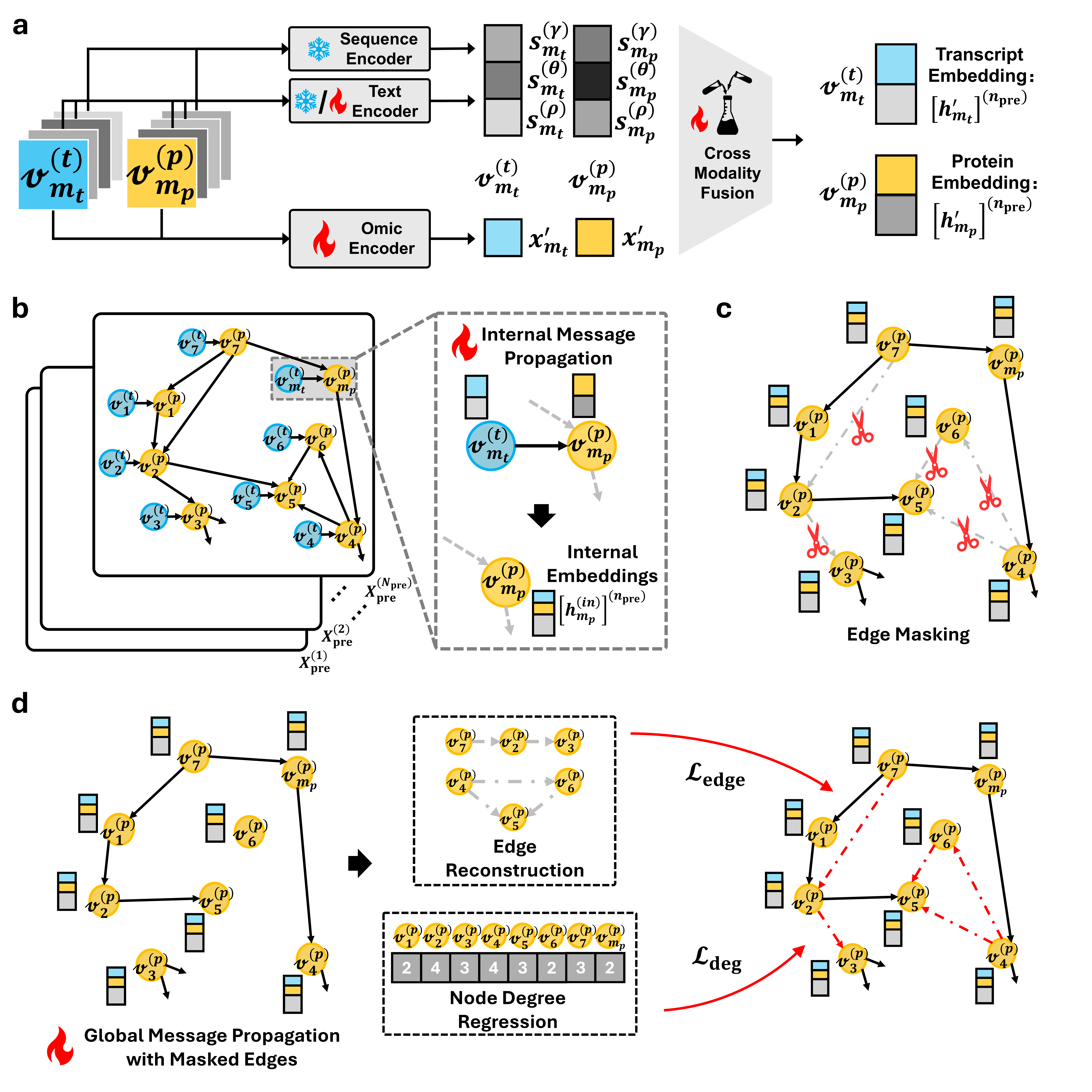}
\caption{\textbf{Overview of CellTOSG-FM Pretraining.}
\textbf{(a)} Cross-modal representation learning: a text encoder and an omic encoder are fused to embed nodes.
\textbf{(b)} Messages propagate within nucleus (internal) and across the TOSG (global) to encode fused biological and textual context for pretraining model.
\textbf{(c)} Mask edges by random sampling.
\textbf{(d)} Global message propagation via masked edges. Edge reconstruction and node-degree prediction are used as pretraining objectives, optimized via edge-reconstruction loss ($\mathcal{L}_{\text{edge}}$) and node degree regression loss ($\mathcal{L}_{\text{deg}}$).
}
\label{fig:celltosg-fm}
\vspace{-0.25in}
\end{figure}

We pretrained the model, $f_{\text{pre}}(\cdot)$, in a self-supervised manner on a subset \(\mathcal{X}_{\mathrm{pre}}\subseteq\mathcal{X}\), without using any metadata from the attribute set \(\mathcal{A}\). As shown in Figure~\ref{fig:celltosg-fm}, sequence information is encoded with DNA-GPT~\cite{zhang2023dnagpt} for RNA sequences (thymine \(T\) substituted by uracil \(U\)) and ProtGPT2~\cite{ferruz2022protgpt2} for protein sequences. These encoders provide high-capacity, transferable representations learned from large-scale genomic and proteomic corpora, and have demonstrated strong generalization on diverse downstream biological tasks. Because transcript and protein entities in BioMedGraphica are stable and reused across samples, we freeze the sequence encoders to eliminate redundant computation, reduce overfitting, and ensure reproducibility across runs; users may substitute alternative sequence language models if desired. A trainable omic encoder then maps numerical omic measurements into the same latent space, and a cross-modality encoder integrates the textual/sequence priors with the quantitative omic evidence to form unified entity representations (Figure~\ref{fig:celltosg-fm}a). This design allows the model to leverage complementary information sources, including semantics and biochemistry from sequences and context-specific variation from omics, within a single representational framework.

Furthermore, we incorporate topological structure by encoding the latent representations with graph encoders. Concretely, messages are propagated to protein entities in two coupled stages: an internal message-passing step that aggregates signals within transcript–protein pairs at the nucleus level, followed by a global propagation step that diffuses information within the cell via the protein–protein interaction topology. The latter is trained with stochastically masked edges (Figure~\ref{fig:celltosg-fm}b–c), optimizing a joint objective that combines edge reconstruction with a degree-regularization term to calibrate node centrality. Consistent, monotonic reductions in the training objective are observed, and held-out edge recovery improves throughout optimization; the degree-oriented auxiliary term further sharpens hub–periphery structure and stabilizes learning (Figure~\ref{fig:celltosg-fm}d). Using an edge-masking ratio of \(10^{-5}\) for the self-supervised objective~\cite{li2023s}, the model reconstructs about 80\% of masked edges, and attains an AUC near 0.85 when it converges  (see Figure~\ref{fig:celltosg-fm-performances}). Notably, these outcomes are achieved when pretraining on \(5\%\) of OmniCellTOSG for \(\mathcal{X}_{\mathrm{pre}}\), underscoring the sample efficiency of the approach and its suitability under limited pretraining budgets. Additional details of the CellTOSG-FM pretraining protocol are provided in Section~\ref{celltosg_fm}.

\subsection{Downstream Tasks Based on CellTOSG-FM}
After pretraining the foundation model CellTOSG-FM, we preserve the model architecture $f_{\text{pre}}(\cdot)$ together with its pretrained parameters $\omega_{\text{pre}}$ as a transferable initialization for downstream adaptation. For each downstream task, we employ the \texttt{CellTOSG\_Loader} to extract and organize the task-specific datasets for fine-tuning. In our downstream tasks, lung adenocarcinoma (LUAD), atrial fibrillation (AF) and  systemic lupus erythematosus (SLE) cohorts are sourced from OmniCellTOSG. We additionally evaluate on an external Alzheimer’s disease (AD) single-cell cohort from the GSE129308 project, which is available across both CELLxGENE and Brain Cell Atlas and was held out from OmniCellTOSG to enable an independent evaluation (see Section~\ref{external-ad-processing} for processing details). In general, the extracted samples/cells are first embedded using the pretrained foundation model $f_{\text{pre}}(\cdot)$, together with the downstream omic and text encoders, to generate integrated feature representations. Subsequently, gene-level information for each sample/cell is projected into a latent embedding space, denoted as $\mathcal{Z}^{(\tau)}$, which serves as the input for task-specific predictors and facilitates efficient adaptation across diverse biological tasks (see Figure~\ref{fig:cell-type-performances}a and Section~\ref{downstream} for details).

\subsubsection{CellTOSG-FM Improves Cell Type Annotation}
\begin{figure}[H]
\vspace{-0.15in}
\centering
\captionsetup{skip=2.5pt} 
\includegraphics[width=0.9\textwidth]{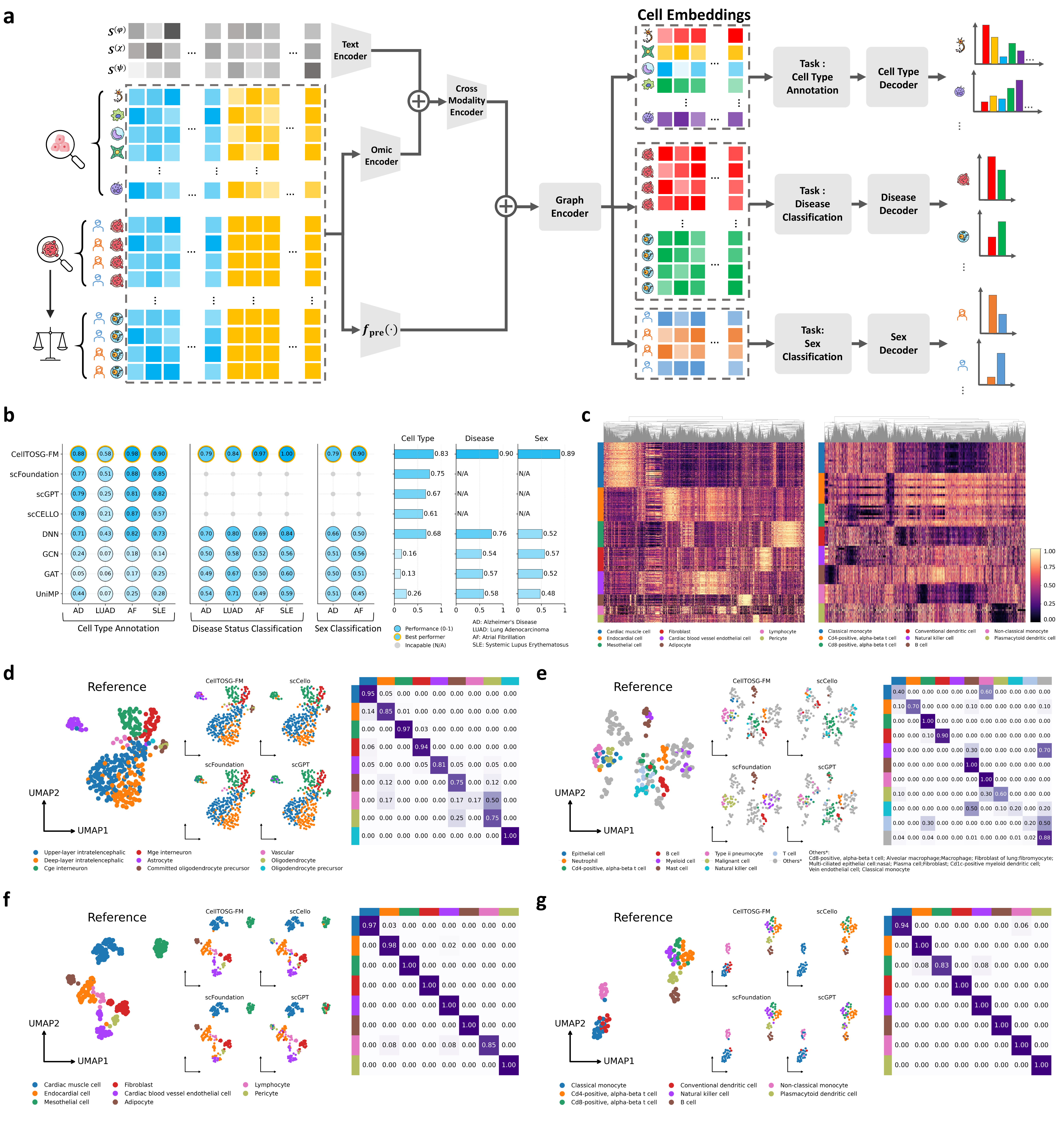}
\caption{\textbf{Experimental results of CellTOSG-FM performance on cell type annotation and cell classification task.}
\textbf{(a)} An illustration of the cell type annotation and cell classification model based on CellTOSG-FM.
\textbf{(b)} Performances on cell type annotations and cell condition classifications (disease vs. normal) on four different diseases (AD, LUAD, AF and SLE), and cell sex classification (male vs. female) on AD and AF.
\textbf{(c)} Cell embeddings learned by the foundation model for cell type annotation in AF and SLE datasets, showing clear separation of cell identities in latent space prior to the classification decoder.
\textbf{(d-g)} Comparison of cell type annotations performances on AD, LUAD, AF and SLE. on UMAP between OmniCellTOSG ground truth and predictions from four models (CellTOSG-FM, scCello, scFoundation, and scGPT). Confusion matrices report the ten most abundant cell types in the sampled dataset, with remaining types grouped as “Others” for visualization while retained as distinct classes for evaluation. 
}
\label{fig:cell-type-performances}
\vspace{-0.25in}
\end{figure}

Using cell-specific embeddings generated from CellTOSG-FM, downstream omic and text encoder and downstream message propagation via GNN layers, the downstream cell-type decoder will be applied to predict the cell-type (Figure~\ref{fig:cell-type-performances}a), and we evaluated annotation performance here on four disease cohorts (AD, LUAD, SLE, and AF). For each cohort, \texttt{CellTOSG\_Loader} was used for cohorts sourced from OmniCellTOSG, while an analogous pipeline was applied to the external AD cohort, to construct balanced downstream datasets by matching the empirical distributions over cell types, thereby mitigating class-imbalance effects. To ensure consistent and comparable evaluation across tasks while reducing computational overhead, we further subsampled approximately 1{,}000 meta-cells for each task-specific dataset. To avoid potential data leakage, samples were partitioned into training and testing sets using donor identity as the splitting criterion, thereby ensuring donor-level independence between the two datasets. The resulting splits typically allocated approximately 20--40\% of samples to the test datasets.

Across all four cohorts, CellTOSG-FM matched or exceeded strong baselines (including DNN~\cite{rumelhart1986learning,hornik1989multilayer}, GCN~\cite{kipf2016semi}, GAT~\cite{velivckovic2017graph}, UniMP~\cite{shi2020masked}, scGPT, scFoundation, and scCELLO) with consistent gains in every disease (see Figure~\ref{fig:cell-type-performances}b and Table~\ref{tab:celltype-performance} for overall performances and Figure~\ref{fig:cell-type-performances}d-g for more details). These improvements indicate that integrating biological textual priors with numerical omic evidence via a cross-modality encoder, together with knowledge-graph–grounded topology in the graph encoder, yields representations that transfer robustly to disease-specific annotation without task-specific tuning. 

Heatmaps derived from the learned cell representations demonstrate clearer block structure and well separated manifold on cell types. Figure~\ref{fig:cell-type-performances}c illustrates the latent embedding spaces for AF and SLE. After restricting the analysis to the 5,000 highest-variance genes and arranging cells by cell type, distinct cell populations occupy clearly separated regions of the latent space in both datasets, demonstrating the effectiveness of CellTOSG-FM in capturing cell-type-specific differences. UMAP projections show that CellTOSG-FM produces compact, well-separated clusters whose boundaries closely align with reference labels, whereas alternative methods exhibit fragmented clusters and label mixing for several closely related types (Figure~\ref{fig:cell-type-performances}d-g). The accompanying confusion matrices display stronger diagonal dominance and fewer systematic off-diagonal errors for CellTOSG-FM. Collectively, these analyses demonstrate that the fused text–omic, graph-aware embeddings produced by CellTOSG-FM deliver higher annotation accuracy and cleaner class separability across diverse disease contexts than single-modality or topology-agnostic baselines.

\subsubsection{CellTOSG-FM Enhances Cell Classification Accuracy}
We extracted disease-specific datasets using \texttt{CellTOSG\_Loader} for OmniCellTOSG cohorts and applying an analogous pipeline to the external AD cohort, balancing cohorts by matching empirical cell-type distributions and thereby mitigating class-imbalance effects. To test whether the fused text–omic, graph-aware representations from CellTOSG-FM support accurate prediction of cell-level attributes, we trained task-specific decoders on top of cell embeddings (Figure~\ref{fig:cell-type-performances}a). Two evaluation settings were considered: disease status (disease vs. normal) in AD, LUAD, SLE and AF sampled dataset; sex classification (male vs. female) in AD and AF sampled dataset. Across all cohorts, CellTOSG-FM achieved the highest accuracies relative to strong baselines (GCN, GAT, DNN, UniMP) (Figure~\ref{fig:cell-type-performances}b and Tables~\ref{tab:disease-performance}-\ref{tab:sex-performance}). These gains were obtained without task-specific architectural changes, indicating that integrating biological textual priors with numerical omic evidence, together with knowledge-graph–grounded topology, yields representations that generalize effectively to diverse attribute-prediction tasks.

To assess model robustness, we examined sex classification within AF dataset, stratified by age groups and major cell types (cardiac muscle cell, cardiac blood vessel endothelial cell, fibroblast cell, adipocyte cell, mesothelial cell, and others) (see Table~\ref{tab:af-sex-age} and Table~\ref{tab:af-sex-celltype}). Heatmaps of the learned gene hyperspace embeddings, restricted to the 5,000 highest variance genes, reveal consistent separation patterns that remain evident when cells are stratified by age group and cell type (Figure~\ref{fig:cell-embed}). Collectively, these analyses indicate that CellTOSG-FM enables accurate and robust predictions across cohorts and biological strata, while the embedding spaces exhibit clear and stable group-wise separation, providing additional structural evidence of model robustness.

\subsection{CellTOSG-FM Is Interpretable to Rank Targets and Signaling Networks}
\begin{figure}[H]
\vspace{-0.15in}
\centering
\captionsetup{skip=2.5pt} 
\includegraphics[width=1.0\textwidth]{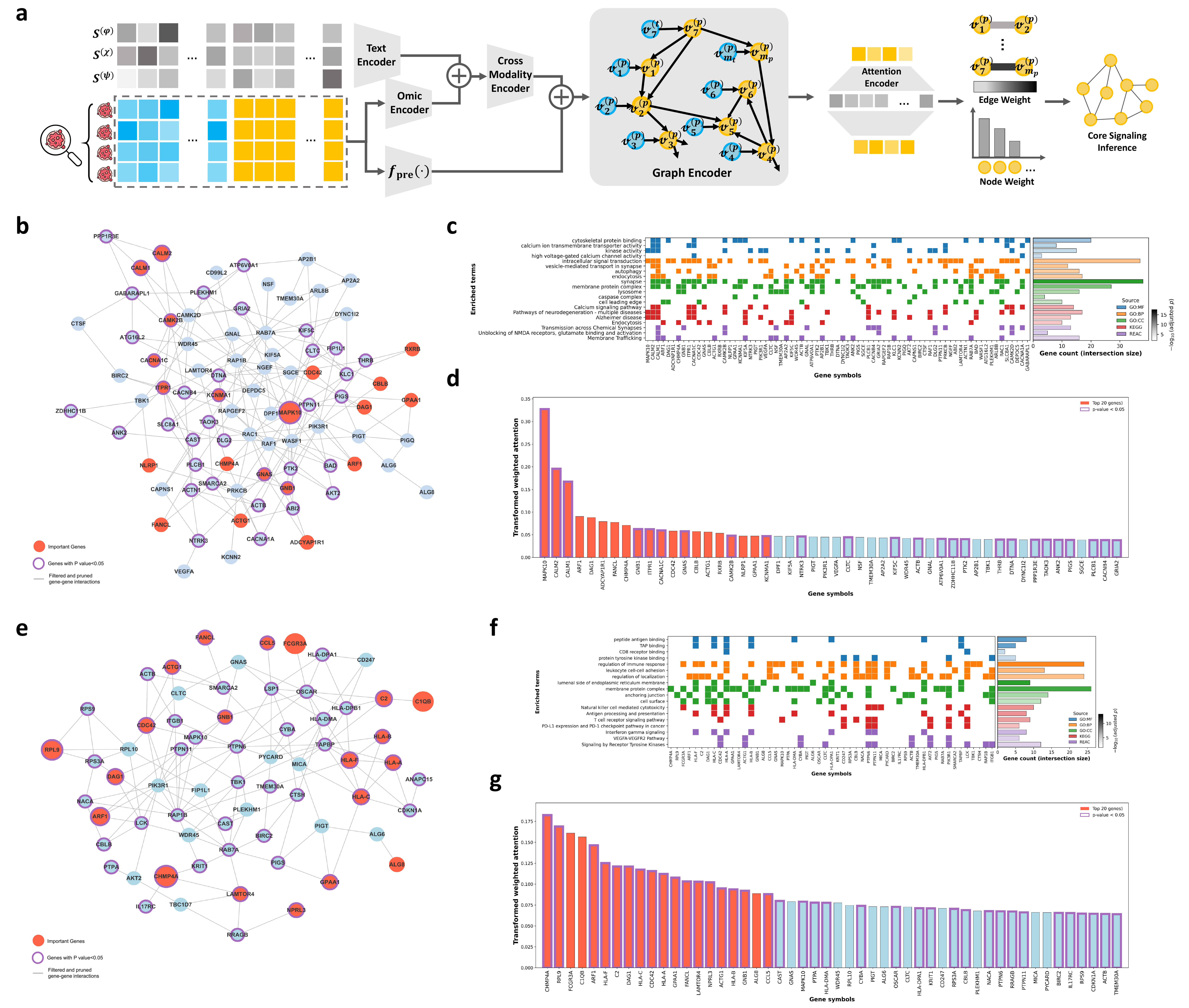}
\caption{\textbf{Target and signaling network inference using CellTOSG-FM.}
\textbf{(a)} Workflow for inferring disease-specific targets and signaling networks from retrieved samples.
\textbf{(b–d)} Analysis results of AD: (b) inferred core disease-relevant signaling subgraph; (c) functional enrichment of genes in the AD core signaling subgraph; (d) ranked normalized gene-importance score distributions.
\textbf{(e–g)} Analysis results of LUAD: (b) inferred core disease-relevant signaling subgraph; (f) functional enrichment of genes in the LUAD core signaling subgraph; (d) ranked normalized gene-importance score distributions.
Purple nodes/bars (in b, d, e, g) mark the top 20 genes by importance; red outlines indicate genes with p value smaller than 0.05 by comparing disease and normal cells.
}
\label{fig:core-infer}
\vspace{-0.25in}
\end{figure}

To illustrate interpretability, signaling targets and pathways were inferred per sample and summarized at the cohort level (Figure~\ref{fig:core-infer}). Latent representations, $\mathcal{Z}^{(\tau)}$, were translated into pairwise affinities between protein entities and then constrained to the protein–protein interaction (PPI) network, yielding edge weights that respect known biology. Node importance was computed by aggregating the strengths of incident edges and modulating by corresponding molecular signals, producing saliency scores that reflect both connectivity and activity. Cohort-level subgraphs were formed from top-ranked genes and pruned to eliminate spurious star-like branches while preserving informative connections, resulting in compact, connected networks aligned to each disease context (see Figure~\ref{fig:core-infer}a and Section~\ref{downstream} for details).

For each sample, directed edge weights between BioMedGraphica nodes were first aggregated and mapped to their corresponding gene names. Edge weights, collected from the attention-based model, sharing the same source–target gene pairs were summed within each sample, reciprocal directions were averaged to obtain undirected edge weights, and edges from each individual sample within the same group were combined by aggregating weights of identical gene pairs across samples (Figure~\ref{fig:core-infer}a). From the attention-based undirected graph, we derived node attention by averaging the attention weights of all edges incident to each node. In parallel, node expression value was derived by selecting, for each gene, the transcript index with the highest mean (min–max normalized) expression across samples, and group-wise average expressions and Mann–Whitney U \(p\)-values of comparing disease and control groups were then calculated for each node. Afterwards, the importance of nodes was ranked by using a node importance score~\citep{zhang2025m3netflow} defined as the product of node attention and averaged expression of the corresponding group. Subsequently, the most important \(\xi=120\) nodes per group were retained. Then, resulting signaling subgraph was refined by keeping only the largest connected component and pruning star-like leaf artifacts while preserving at least \(\epsilon=3\) leaves per node, preferentially retaining those with significant \(p\)-values or otherwise the highest-weight neighbors~\cite{zhang2025mosgraphflow}.

Fig~\ref{fig:core-infer}a summarizes the pipeline from retrieved disease cohorts to targets and signaling networks subgraphs. Fig~\ref{fig:core-infer}b–d show results for Alzheimer’s disease, where the inferred targets and signaling networks network recovers a compact set of connected signaling modules (red) with multiple statistically supported genes (purple outlines; \(p<0.05\)), and functional enrichment recapitulates key neurodegenerative processes (Figure~\ref{fig:core-infer}c–d). Enrichment analysis highlights synapse-associated and cytoskeleton-related programs, including synapse and cytoskeletal protein binding, consistent with extensive evidence that early AD pathogenesis involves synaptic vulnerability and disrupted actin/cytoskeletal regulation~\cite{penzes2011impaired, pelucchi2020dendritic, jiang2025cytoskeletal}. Calcium-related molecular functions and pathways are also well represented, including calcium ion transmembrane transporter activity, high voltage-gated calcium channel activity, and Calcium signaling pathway, supporting the long-standing Ca\(^{2+}\) dysregulation framework in AD and implicating VGCC/NMDAR-linked signaling as a mechanistic contributor to synaptic failure and neurotoxicity~\cite{stutzmann2005calcium, ghosh2015calcium, crossley2023modulation}. Consistently, synaptic transmission terms (Transmission across Chemical Synapses and Unblocking of NMDA receptors, glutamate binding and activation) align with prior work linking aberrant glutamatergic/NMDAR signaling to AD synaptic dysfunction~\cite{liu2019role, lanskey2025alzheimer}. Trafficking-related biology is strongly supported by enrichment of endocytosis, vesicle-mediated transport in synapse, Endocytosis, and Membrane Trafficking, supported by evidence that endocytosis and vesicle recycling defects are early and mechanistically relevant to AD~\cite{kelly2007beta, perdigao2020intracellular}. Moreover, enrichment of autophagy and lysosome aligns with reports of autophagy--lysosome dysfunction in AD~\cite{zhang2022impairment, nixon2024mechanisms}. Finally, kinase-centric and signaling-related terms (kinase activity and intracellular signal transduction) are consistent with evidence that dysregulated neuronal signaling cascades contribute to AD progression ~\cite{li2023targeting, wu2025protein}, suggesting that the inferred signaling pathways form a compact, connected network consistent with reference neurodegeneration/AD pathway frameworks.

Figure~\ref{fig:core-infer}e--g show the inferred LUAD targets and signaling networks network and its functional enrichment. The enrichment results are dominated by antigen presentation and immune effector programs, with additional checkpoint and growth/angiogenesis signaling. Enrichment of antigen processing/presentation pathways and functional categories (e.g., peptide antigen binding, TAP binding, and Antigen processing and presentation) is consistent with established mechanisms by which altered HLA/APM function shapes tumor immune evasion and immunotherapy responsiveness in lung cancer~\cite{thompson2020gene, kubo2024assessment}. In parallel, enrichment for cytotoxic immune pathways (Natural killer cell mediated cytotoxicity and T cell receptor signaling) aligns with the central role of NK/T-cell–mediated tumor surveillance and effector function within the LUAD tumor microenvironment~\cite{zeng2021natural}. Checkpoint and cytokine signaling (PD-L1 expression and PD-1 checkpoint pathway in cancer together with Interferon gamma signaling) aligns with prior evidence that IFN-$\gamma$–driven activation programs can also promote adaptive immune resistance via PD-L1 upregulation~\cite{mandai2016dual, lingling2020molecular, hirayama2023regulation}. Finally, enrichment of Signaling by Receptor Tyrosine Kinases and the VEGFA--VEGFR2 Pathway is compatible with canonical growth and angiogenesis programs in NSCLC, which are also known to interact with immune regulation in the tumor microenvironment~\cite{ntellas2020old, ghalehbandi2023role, lee2025vascular}.

Across diseases, the importance score distributions highlight biologically coherent modules, suggesting that the attention-derived, PPI-constrained geometry together with feature-scaled importance scores helps capture disease-associated signaling patterns that are compact, connected, and functionally interpretable.



\section{Dicussion}\label{discussion}
Tissue-level and single-cell omic resources are being generated at unprecedented scale to interrogate disease pathogenesis—the core of precision medicine. Graph neural networks (GNNs) have been widely used to integrate molecular measurements with interaction knowledge for target identification and pathway inference~\cite{wang2021mogonet,zhang2025m3netflow,dong2023highly,zhang2024using}. Nevertheless, despite strong predictive performance, prevailing graph-based approaches that operate on numeric, expression-centric signaling graphs capture only part of the scientific discovery workflow: they often underutilize the rich, human-interpretable priors encoded in biological text and curated knowledge bases. To address this gap, this work introduces a three-part ecosystem that unifies data, tooling, and modeling. First, \textbf{OmniCellTOSG} is a large-scale single-cell text–omic signaling graph resource whose \textbf{Text–Omic Signaling Graphs (TOSGs)} couple text-attributed biological knowledge with numerical gene/protein abundance, enabling graph-structured decoding of cellular signaling across tissues, diseases, ages, and conditions. Second, \texttt{\textbf{CellTOSG\_Loader}} provides a NumPy-ready query–load–balance pipeline that constructs stratified, unbiased cohorts across user-specified facets (e.g., cell type, tissue, disease, data source, age/sex), mitigates class imbalance, and applies batch correction, standardizing ingestion and ensuring fair, repeatable experimentation. Third, the \textbf{CellTOSG Foundation Model (CellTOSG-FM)} integrates a graph-language architecture that jointly encodes biological textual priors, quantitative omic measurements, and signaling topology over TOSGs. By enabling message passing on knowledge-grounded graphs while aligning cross-modal representations, the model learns structure-aware embeddings that support downstream tasks including cell-type annotation, cell-attribute classification, and signaling inference with interpretable subgraph rationales.

These properties position TOSGs as a natural substrate for foundation-model training, enabling the learning of broadly transferable model of cellular signaling. Pretraining \textbf{CellTOSG-FM} on massive, heterogeneous TOSG corpora from \textbf{OmniCellTOSG} via self-supervised learning, which leverages edge-masked reconstruction that emphasizes signaling network structure, yields broadly transferable model of signaling patterns and provides robust bases for task-specific adaptation, outperforming disease- or dataset-specific pipelines that risk bias and overfitting. The \textbf{OmniCellTOSG} dataset is openly available in a PyTorch-friendly format, lowering barriers to reproducible benchmarking and catalyzing community development of graph language foundation models for precision medicine over cellular systems. Together, \textbf{OmniCellTOSG}, \texttt{\textbf{CellTOSG\_Loader}}, and \textbf{CellTOSG-FM} establishes a scalable, mechanism-oriented framework for cell type annotation, disease classification, subtype delineation, and targets and signaling networks graph inference. Ongoing curation continues to expand its coverage across diseases, tissues, sex, age, and diverse experimental conditions, enabling improved interrogation of complex signaling programs and the prioritization of actionable perturbations, including candidate drugs and rational combinations that target dysfunctional nodes and pathways.

\section{Methods}\label{method}
\subsection{Data Collection and Preprocessing} 
\label{data_processing}
The dataset was compiled primarily from three large sources, with additional cohorts added to broaden tissue and disease coverage (collection procedures in Section~\ref{data-download}). From CellxGene, we obtained over 71 million single cells/nuclei across 65 human tissues and 125 disease studies in H5AD AnnData format~\cite{megill2021cellxgene,czi2025cz}; from the Brain Cell Atlas, over 7 million human brain single cells spanning 21 disease types~\cite{chen2024brain,yao2023high}; and from GEO, four studies contributing over 650{,}000 cells to fill underrepresented indications. We further integrated the Hepatitis Atlas to include hepatitis~C virus infection data (over 7{,}000 cells) and the Human Cell Atlas to expand pancreas coverage with over 98{,}000 cells and three additional disease conditions. All datasets were converted to a unified H5AD schema to support a standardized preprocessing workflow. The resulting preprocessed resource comprises over 79 million high-quality cells covering 762 cell types, with samples organized and split by source, coarse- and fine-grained tissue labels, disease, and suspension type.

To mitigate the inherent sparsity and noise in sc/snRNA-seq data, we adopt a meta-cell strategy based on the SEACells algorithm\cite{persad2023seacells}. Our approach is designed to ensure consistency across datasets from diverse sources by employing uniform preprocessing, feature selection, and dimensionality reduction procedures before meta-cell aggregation. Let the raw data be represented by $\mathcal{X}^{(\alpha)}=\{X^{(\alpha)}_{1},X^{(\alpha)}_{2},\cdots,X^{(\alpha)}_{n_0},\cdots, X^{(\alpha)}_{N_0}\}$, where $X^{(\alpha)}_{n_0}\in{\mathbb{R}^{M_{0}}}$ denotes the cell, and $N_0$ is the number of cells collected from various data resources and $M_0$ is the number of elements in transcript entity set $\mathcal{T}=\{T_1,T_2,\cdots,T_{m_0},\cdots,T_{M_0}\}$. For computational demands, raw data files (stored in H5AD format) are partitioned into subsets of no more than 50,000 cells. For datasets requiring normalization, we first apply total count normalization by scaling UMI counts of each cell to a fixed total of 10,000, followed by a log1p transformation to stabilize variance. In addition, Uniform feature selection is performed by identifying the top 1,500 highly variable genes from each dataset. We then apply Principal Component Analysis (PCA\cite{abdi2010principal}) with 50 components to reduce dimensionality while preserving essential variance. Based on the PCA-reduced features, a K-Nearest Neighbor (KNN\cite{peterson2009k}) graph is constructed to maintain the underlying structural relationships among cells. Meta-cell generation is performed using the SEACells algorithm. With a fixed aggregation size of $N$ cells per meta cell, SEACells first measures cell-to-cell similarity and then decomposes the resulting structure via archetypal analysis. Cells near the convex hulls of the data distribution are grouped together, yielding a new set of meta cells denoted by $\mathcal{X}^{(\beta)}=\{X^{(\beta)}_1,X^{(\beta)}_2,\cdots,X^{(\beta)}_n,\cdots, X^{(\beta)}_{N}\}$, where $X^{(\beta)}_{n}\in{\mathbb{R}^{M_0}}$ represents a meta-cell.

Correspondingly, the associated attributes (e.g., sex, cell\_type, development\_stage, tissue, disease\_status, etc.) for the meta cells are computed by aggregating the raw cell attributes through majority voting, resulting with $\mathcal{A}^{(\beta)}$ (see \textbf{Figure~\ref{fig:architecture}a}). After all meta-cells are constructed, we group the data by source and tissue (coarse-grained), read the corresponding meta-cell H5AD files, and map gene identifiers to the BioMedGraphica framework (see Section~\ref{tosg_generation}). The expanded expression matrices are then serialized into NumPy shards of 10{,}000 samples per file. For each sample, we record the relative path of the NumPy matrix (matrix\_file\_path) and the corresponding row index in the matrix file (matrix\_row\_idx); these pointers are stored together with the H5AD .obs fields and saved in CSV format for downstream processing. Because the datasets originate from diverse sources, the nomenclature of attributes such as cell type, disease, development stage, and sex varies substantially.To standardize cell type, we built a mapping pipeline using the unique cell type values extracted from all datasets and a cell type mapping table (CMT) based on Cell Ontology (CL) database. All synonym fields in the cell type mapping table were expanded to generate a candidate dictionary linking each synonym to its CMT ID and the corresponding CL label. Before matching, anchor rules were defined to handle specific terms and generic placeholders such as “unknown” or “unclassified” were ignored. For every original cell type, exact matching was first attempted; if not found, fuzzy matching (token-sort ratio) was applied to compute the best-scoring candidate. The resulting pairs contained the original term, the matched CMT term, CMT ID, CL label, and the matching score. Terms with scores lower than 100 were manually reviewed and corrected, producing a curated mapping that unified all cell type under the Cell Ontology standard. The same procedure was used for disease terms to obtain BMG disease identifiers.

The development stage values were normalized by converting free-text descriptions into approximate numeric ages (in years) using regular-expression parsing of units such as years, months, weeks, days, and Carnegie stages, followed by categorization into MeSH-based age groups (e.g., infant, child, young adult, middle aged)\cite{MeSH_Age_Groups}. Each entry was also assigned a coarse birth phase label (pre-birth, post-birth, or unknown). For sex, all terms were normalized through direct mapping (e.g., f for female, m for male), with unrecognized or empty values set to unknown. Finally, by integrating the metadata with the curated mapping results, we obtain the standardized attribute set $\mathcal{A}=\{a_i\}_{i=1}^m$ (Table~\ref{tab:attributes}). This set serves as retrieval metadata for \texttt{CellTOSG\_Loader} to locate and extract the corresponding cells (Section~\ref{tosg_loader}).

\subsection{OmniCellTOSG Generation}
\label{tosg_generation}
\noindent With the preprocessed single-cell transcriptomic dataset denoted as \(\mathcal{X}^{(\beta)} \in \mathbb{R}^{N \times M_0}\), we integrate it into the BioMedGraphica framework together with the gene-regulatory network. Using the mapping table, the \(M_0\) transcript features are mapped to \(M_t\) transcript entities. Specifically, each transcript element in the set \(\mathcal{T}\) is mapped and extended to the transcript-entity set \(\mathcal{V}^{(t)}=\{v^{(t)}_{1},\,v^{(t)}_{2},\,\ldots,\,v^{(t)}_{M_t}\}\). By linking transcript nodes within the network to the protein–protein interaction (PPI) graph, proteins are treated as virtual nodes, yielding the additional entity set \(\mathcal{V}^{(p)}\). The overall entity set is \(\mathcal{V}=\{\mathcal{V}^{(t)},\,\mathcal{V}^{(p)}\}\), with \(\lvert \mathcal{V}\rvert=\lvert \mathcal{V}^{(t)}\rvert+\lvert \mathcal{V}^{(p)}\rvert=M_t+M_p=M\). Likewise, the feature set \(\mathcal{X}=\{\mathcal{X}^{(t)},\,\mathcal{X}^{(p)}\}\) is generated, where \(\mathcal{X}\in\mathbb{R}^{N\times M}\), \(\mathcal{X}^{(t)}\in\mathbb{R}^{N\times M_t}\), and \(\mathcal{X}^{(p)}\in\mathbb{R}^{N\times M_p}\) correspond to the transcriptomic and proteomic feature sets, respectively.

From the perspective of single cell side, the multi-omics $\mathcal{X}$ can be decomposed as $\{X_1,X_2,\cdots,X_n,\cdots,X_N\}$, where each sample $X_n$ resides in $\mathbb{R}^M$. Additionally, the cell label matrices set $\mathcal{Y}$, and given that the cell label set are consistent with label for meta cells, $\mathcal{Y}^{(\beta)}$. Beyond transcriptomic features and virtual proteomic features, an auxiliary node textual information dataset, $\mathcal{S} = \{S^{(\varphi)}, S^{(\chi)},S^{(\psi)}\}$, is incorporated. Each of those entity textual information correpsonds to the node in entity set $\mathcal{V}$.  The $S^{(\varphi)}=[s^{(\varphi)}_{1},s^{(\varphi)}_{2},\cdots,s^{(\varphi)}_{m},{\cdots},s^{(\varphi)}_{M}]$, representing the entity names (e.g., HGNC symbol, Ensembl ID), $S^{(\chi)}=[s^{(\chi)}_{1},s^{(\chi)}_{2},\cdots,s^{(\chi)}_{m},\cdots,s^{(\chi)}_{M}]$, representing the entity textual descriptions (e.g., Uniprot protein description), and $S^{(\psi)}=[s^{(\psi)}_{1},s^{(\psi)}_{2},\cdots,s^{(\psi)}_{m},\cdots,s^{(\psi)}_{M}]$, representing biochemical information (i.e., RNA sequences or protein sequences). Therefore, for any entity, $v_m$, it has the textual information set $s_m=\{s^{(\varphi)}_{m},s^{(\chi)}_{m},s^{(\psi)}_{m}\}$. And the entity textual information dataset, $\mathcal{S}$, enhances the graph's expressivity, facilitating the generation of a textual-attributed transcriptomic signaling knowledge graph. 

Afterwards, to construct the text–omic signaling graph \(\mathcal{G} = (\mathcal{V}, \mathcal{E})\), we identify relations (edges) between entities. As noted above, the vertex set is \(\mathcal{V} = \{\mathcal{V}^{(t)}, \mathcal{V}^{(p)}\}\). We consider two relation types: internal signaling and PPI-based gene-regulatory signaling. Accordingly, the graph decomposes into the internal-signaling subgraph \(\mathcal{G}^{(\mathrm{in})} = (\mathcal{V}^{(\mathrm{in})}, \mathcal{E}^{(\mathrm{in})})\), which captures the molecular flow from transcripts to proteins, and the PPI-regulatory subgraph \(\mathcal{G}^{(\mathrm{PPI})} = (\mathcal{V}^{(\mathrm{PPI})}, \mathcal{E}^{(\mathrm{PPI})})\), which captures protein–protein interactions, with the overall edge set \(\mathcal{E}=\mathcal{E}^{(\mathrm{in})}\cup\mathcal{E}^{(\mathrm{PPI})}\). By construction, \(\mathcal{V}^{(\mathrm{in})}=\mathcal{V}\) with \(\lvert \mathcal{V}^{(\mathrm{in})}\rvert = M = M_t + M_p\), while \(\mathcal{V}^{(\mathrm{PPI})}=\mathcal{V}^{(p)}\). 

Overall, the pipeline condenses \(79{,}195{,}364\) raw cells into \(N=395{,}317\) meta-cells and aligns \(M=533{,}458\) molecular entities (transcript/protein nodes) enriched with textual and topological information with internal signaling edges with $\lvert\mathcal{E}^{(\text{in})}\rvert=152,585$ and PPI-regulatory subgraph with $\lvert\mathcal{E}^{(\mathrm{PPI})}\rvert=16,484,820$. Building on these components, we fuse preprocessed single-cell transcriptomic profiles with prior gene–regulatory and signaling knowledge to construct TOSGs, and we release the dataset \(\mathcal{D}=(\mathcal{X},\mathcal{A},\mathcal{S},\mathcal{E})\). The resulting TOSGs provide a unified, graph-structured substrate for foundation-model pretraining and downstream tasks by coupling numeric omics measurements with textual and topological knowledge, thereby enabling structure-aware learning and interpretable signaling analysis. A comprehensive summary of organ and disease coverage is provided in Table~\ref{tab:data_overview}.\\

\subsection{\texttt{CellTOSG\_Loader} Package} \label{tosg_loader}
To enable scalable access, the feature matrix \(\mathcal{X}\in\mathbb{R}^{N\times M}\) in \textbf{OmniCellTOSG} is partitioned row-wise into fixed-size NumPy shards, each stored as an \texttt{x.npy} file with lightweight metadata recording global row indices. After downloading the dataset to a local \texttt{root}, users employ \texttt{CellTOSG\_Loader} (Appendix~\ref{celltosg_loader}), which discovers the relevant shards and materializes only the requested subset, thereby avoiding full-matrix loads. Cohorts are defined via standardized metadata filters \(\texttt{conditions}\) (e.g., \{\texttt{tissue\_general}: brain, \texttt{disease\_name}: Alzheimer’s Disease\}) expressed over the attribute set \(\mathcal{A}\); the supervised objective and target field are designated by \(\texttt{task}\) and \(\texttt{label\_column}\), respectively. Given these inputs, the loader deterministically translates user arguments into a formal query, extracts the feasible subset from \(\mathcal{X}\), and optionally applies subsampling (to optimize memory usage), class balancing, and batch correction via ComBat--seq~\cite{zhang2020combat,behdenna2023pycombat}. At the core of retrieval is a two-phase Stratified Retrieval Algorithm (SRA) tailored to \(\mathcal{X}\) and its named attributes \(\mathcal{A}\) as meta data. In \textbf{Phase I (query-constrained extraction)}, a user’s conjunctive query yields the subset \(R(q)\) by enforcing that all specified attribute constraints hold. In \textbf{Phase II (task-aware balancing)}, a task configuration specifies the balance label and control value, the exact-match covariates, and an ordered age–stage key; cases are taken from \(R(q)\) (non-control label), controls are drawn by reapplying the same filters with the label fixed to the control value, and key-stratified matching is performed: non-age covariates must match exactly within each stratum, while age differences are bounded by a tolerance \(\delta\) along the ordered age–stage axis (with optional upsampling and discarding infeasible strata). The outcome is a stratified cohort \(\mathcal{X}_R\) whose non-stage covariates are balanced by construction and whose age-stage offsets satisfy \(d_{k^\star}\le\delta\). For cell-type annotation, balancing is disabled and rare types (\(<\mu\) samples) may be upsampled for training stability. The procedure simultaneously returns the label set \(Y_R\) aligned with \(\mathcal{X}_R\). The Stratified Retrieval Algorithm is described in detail below.

In the Query-Constrained Extraction phase, we let \(\mathcal{X}\) be the set of samples and \(\mathcal{A}\) the set of attributes. For each \(a\in\mathcal{A}\) with value space \(\Sigma_a\), define the attribute–evaluation map
\(u_a:\mathcal{X}\to\Sigma_a,\ x\mapsto u_a(x)\).
Equivalently, collect these into a single evaluation map
\(u:\mathcal{X}\times\mathcal{A}\to\bigcup_{a\in\mathcal{A}}\Sigma_a\) via \(u(x,a):=u_a(x)\).
A user query is a finite set of attribute constraints
\(q=\{(a,V_a)\}_{a\in\operatorname{dom}(q)}\) with nonempty admissible sets \(V_a\subseteq\Sigma_a\) and
\(\operatorname{dom}(q)=\{\,a\in\mathcal{A}:V_a\neq\varnothing\,\}\), interpreted by the conjunctive predicate
\begin{equation}
  Q_q(x) \;=\; \bigwedge_{(a,V_a)\in q}\, \mathcal{I}\{\,u_a(x)\in V_a\,\},
\end{equation}
where \(\mathcal{I}\{\cdot\}\in\{0,1\}\) equals \(1\) when the statement is true and \(0\) otherwise.
The selection induced by the query is the feasible set
\begin{equation}
  R(q)\;:=\;\{\,x\in \mathcal{X}:Q_q(x)=1\,\}
  \;=\;\bigcap_{(a,V_a)\in q} u_a^{-1}(V_a),
\end{equation}
i.e., the set of samples that satisfy all attribute-wise constraints simultaneously.

During Phase II of Task-Aware Balancing,  we Let the task configuration be \(\Lambda(\lambda)=(b,b_0,K,k^\star)\), where \(b\in\mathcal{A}\) is the balance field with control value \(b_0\in\Sigma_b\), and \(K=(k_1,\dots,k_r)\subseteq\mathcal{A}\) is the ordered tuple of match keys with designated age–stage key \(k^\star=k_{j^\star}\) for some index \(1\le j^\star\le r\).
Define the query with any constraint on \(b\) removed by
\begin{equation}
  q_{-b} \;=\; \{(a,V_a)\in q : a\neq b\}.
\end{equation}
The control pool applies the same non-\(b\) filters and overwrites \(b\) to its control value:
\begin{equation}
  \mathrm{NM}(\lambda,q) \;=\; \{\,x\in \mathcal{X} : u_b(x)=b_0 \;\wedge\; Q_{q_{-b}}(x)=1 \,\}.
\end{equation}
The case set is drawn from the query subset:
\begin{equation}
  \mathrm{CA} \;=\; \{\,x\in R(q) : u_b(x)\neq b_0 \,\}.
\end{equation}
Let \(\Sigma_K := \Sigma_{k_1}\times\cdots\times\Sigma_{k_r}\) and define the key map
\begin{equation}
  \kappa_K: \mathcal{X}\to \Sigma_K,\qquad \kappa_K(x)=(u_{k_1}(x),\dots,u_{k_r}(x)).
\end{equation}
For any key tuple \(\kappa\in\Sigma_K\), define the strata
\begin{equation}
  \mathrm{CA}_\kappa \;=\; \{x\in \mathrm{CA} : \kappa_K(x)=\kappa\}, 
  \qquad
  \mathrm{NM}_\kappa \;=\; \{x\in \mathrm{NM}(\lambda,q) : \kappa_K(x)=\kappa\}.
\end{equation}
Endow \((\Sigma_{k^\star},\preceq)\) with a rank map \(\rho:\Sigma_{k^\star}\to\{0,1,\dots,L\}\) and induced distance
\begin{equation}
  d_{k^\star}(e,e') \coloneqq \lvert \rho(e)-\rho(e') \rvert .
\end{equation}
Given a tolerance \(\delta\ge 0\), the offset-admissible control pool for stratum \(\kappa\) is
\begin{equation}
  \mathcal{N}^{(\delta)}_\kappa
  = \Bigl\{\,x\in \mathrm{NM}(\lambda,q):\ u_{k_i}(x)=\kappa_i\ \forall\, i\neq j^\star,\ 
  d_{k^\star}\!\bigl(u_{k^\star}(x),\kappa_{j^\star}\bigr)\le \delta \,\Bigr\}.
\end{equation}
Matching proceeds per stratum by first taking exact-stage controls (\(d_{k^\star}=0\)); if insufficient, progressively admitting offsets \(1,2,\dots,\delta\) according to the stage order, sampling without replacement within each offset layer, and finally (if enabled) upsampling with replacement from the collected controls to meet the stratum size. Strata with no admissible controls are discarded. Let \(\mathrm{CA}'_\kappa\) and \(\mathrm{NM}'_\kappa\) denote the retained case rows and matched controls for successful strata. The balanced output is
\begin{equation}
  \mathcal{X}_{R}=\bigcup_{\kappa\in\mathcal{K}^\ast}\!\bigl(\mathrm{CA}'_\kappa \cup \mathrm{NM}'_\kappa\bigr),
  \qquad
  \text{with}\ \ \lvert \mathrm{CA}'_\kappa\rvert=\lvert \mathrm{NM}'_\kappa\rvert\quad\forall\,\kappa\in\mathcal{K}^\ast,
\end{equation}
where \(\mathcal{K}^\ast\) is the set of strata that achieved feasible matching. Within each retained stratum, all non-stage keys in \(K\) match exactly by construction, and stage differences satisfy \(d_{k^\star}\le \delta\). Finally, the task-specific labels \(Y_R\) are obtained by restricting the label map \(\ell_b:\mathcal{X}\!\to\!\Sigma_b\) to the returned cohort, i.e., \(Y_R=\{\ell_b(x):x\in\mathcal{X}_R\}\), which is one-to-one aligned with the rows. The details of the algorithm can be checked in the Appendix~\ref{sra_algo}.

\subsection{Graph Language Foundation Model} \label{celltosg_fm}
\subsubsection{CellTOSG-FM pretraining} \label{pretrain}
Given the integrated text--omic signaling graph dataset \(\mathcal{D}\), which comprises a single-cell text--omic signaling graph \(\mathcal{G}=(\mathcal{V},\mathcal{E})\) together with text--omic feature sets \(\mathcal{X}\) and \(\mathcal{S}\), we construct a self-supervised pretraining task by sampling a subset \(\mathcal{X}_{\mathrm{pre}}\subseteq\mathcal{X}\). For edge-masking pretraining, we draw an edge mask set \(\mathcal{E}_{\mathrm{mask}}\sim\mathrm{Bernoulli}(p)\) over the protein–protein interaction subset \(\mathcal{E}^{(\mathrm{PPI})}\), where \(p\in(0,1)\) denotes the masking ratio used to occlude signaling flow along PPI edges. The foundation model is then pretrained via
\begin{equation}
    \mathcal{H}=f_{\text{pre}}(\mathcal{\mathcal{X}_{\text{pre}},S,E, E_{\text{mask}}})
\end{equation}
, where $\mathcal{H}\in{\mathbb{R}^{N\times{M}\times{d}}}$ is the entity embeddings, and $f_{\text{pre}}(\cdot)$ is the pre-trained foundation model. In details, to merge the text-omics feature sets $\mathcal{X,S}$ into unified entity embeddings, bi-encoder framework was leveraged by
\begin{equation}
    \mathcal{X'}=\text{ENC}^{\text{pre}}_{\text{Omic}}(\mathcal{X}_{\text{pre}})
\end{equation}
\begin{equation}
    \mathcal{S'}=\text{ENC}_{\text{Text}}(\mathcal{S})
\end{equation}
\begin{equation}
    \mathcal{H'}=\text{ENC}^{\text{pre}}_{\text{Cross}}(\mathcal{X', S'})
\end{equation}
, where the $\text{ENC}^{\text{pre}}_{\text{Omic}}(\cdot)$ is the linear transformation and $\mathcal{X'}\in{\mathbb{R}^{N\times{M}\times{d'}}}$  the $\text{ENC}_{\text{Text}}(\cdot)$ can be BERT-based or other language models (LMs) and $\mathcal{S'}=\{S^{(\gamma)},S^{(\theta)},S^{(\rho)}\}$, where $S^{(\gamma)}\in{\mathbb{R}^{M\times{d'}}}$, $S^{(\theta)}\in{\mathbb{R}^{M\times{d'}}}$, $S^{(\rho)}\in{\mathbb{R}^{M\times{d'}}}$ are encoded as entity name, textual description and biochemical embeddings. The $\text{ENC}^{\text{pre}}_{\text{Cross}}(\cdot)$ will fuse the omic embeddings and the textual embeddings to generate $\mathcal{H'}\in{\mathbb{R}^{N\times{M}\times{d'}}}$. 

Afterwards, the internal signaling will be propagated by using graph encoder with 
\begin{equation}
    \mathcal{H}^{(\text{in})}=\text{GNN}_{\text{in}}^{\text{pre}}(\mathcal{H'},\mathcal{E}^{\text{(in)}})
\end{equation}
, where $\text{GNN}_{\text{in}}^{\text{pre}}(\cdot)$ uses message propagation via GNN layers and $\mathcal{H}^{(\text{in})}\in{\mathbb{R}^{N\times{M}\times{d}}}$. Finally, with the prepared entity embedding, the foundation model will be pretrained by masking nodes with
\begin{equation}
    \mathcal{H}=\text{GNN}_{\text{global}}^{\text{pre}}\left(\text{MLP}_{\text{pre}}(\mathcal{H}^{(\text{in})}),\mathcal{E}^{\text{(PPI)}},\mathcal{E}_{\text{mask}}\right)
\end{equation}
, where $\text{GNN}_{\text{global}}^{\text{pre}}(\cdot)$ denotes global signaling message propagation via GNN layers and $\text{MLP}_{\text{pre}}(\cdot)$ represents the multilayer perceptron to embed the node features before the final stage of pretraining. 

To update the pretraining model parameters, we adopt a reconstruction objective. Specifically, we define the set of visible edges as \(\mathcal{E}_{\mathrm{vis}}=\mathcal{E}^{(\mathrm{PPI})}\setminus \mathcal{E}_{\mathrm{mask}}\). The structural decoder \(u_{\omega}\) with parameters \(\omega\) outputs the probability of an edge between nodes \(m_i\) and \(m_j\) as
\begin{equation}
u_{\omega}\!\big(h_{m_i}^{(n)},h_{m_j}^{(n)}\big)
= \sigma\!\left(\mathrm{MLP}_{\omega}\!\big(h_{m_i}^{(n)}\!\odot h_{m_j}^{(n)}\big)\right),
\end{equation}
where \(h_{m_i}^{(n)},h_{m_j}^{(n)}\in\mathbb{R}^{d}\) are the global node embeddings for nodes \(m_i\) and \(m_j\) in sample \(n\); \(\mathrm{MLP}\) denotes a multilayer perceptron and \(\odot\) the element-wise product. In parallel, the degree decoder \(v_{\phi}\) with parameters \(\phi\) predicts the node degree via
\begin{equation}
v_{\phi}\!\big(h_m^{(n)}\big) \;=\; \mathrm{MLP}_{\phi}\!\big(h_m^{(n)}\big)
\end{equation}

We compute the pretraining reconstruction loss as the sum of an edge-reconstruction term and a degree-reconstruction term, with \(\mathcal{E}^{+}=\mathcal{E}_{\mathrm{vis}}\) and a set of sampled negative edges \(\mathcal{E}^{-}\) drawn from pairs not in \(\mathcal{E}^{\text{PPI}}\). The binary cross-entropy edge losses for sample \(n\) are
\begin{equation}
\mathcal{L}_{\mathrm{+}}^{(n)}=\frac{1}{\lvert \mathcal{E}^{+} \rvert}\!
  \sum_{(m_i,m_j)\in \mathcal{E}^{+}} \log u_{\omega}\!\big(h_{m_i}^{(n)},h_{m_j}^{(n)}\big)
\end{equation}
\begin{equation}
\mathcal{L}_{\mathrm{-}}^{(n)}=\frac{1}{\lvert \mathcal{E}^{-} \rvert}\!
  \sum_{(m_i,m_j)\in \mathcal{E}^{-}} \log\!\Big(1 - u_{\omega}\!\big(h_{m_i}^{(n)},h_{m_j}^{(n)}\big)\Big)
\end{equation}
\begin{equation}
\mathcal{L}_{\mathrm{edge}}^{(n)}
= -(\mathcal{L}_{\mathrm{+}}^{(n)}+\mathcal{L}_{\mathrm{-}}^{(n)})
\end{equation}
and the degree-reconstruction loss is
\begin{equation}
\mathcal{L}_{\mathrm{deg}}^{(n)}
= \frac{1}{\lvert \mathcal{V} \rvert} \sum_{p \in \mathcal{V}}
  \big\|\, v_{\phi}\!\big(h_{m}^{(n)}\big) - \operatorname{deg}\!\big(m^{(n)}\big) \,\big\|^{2}
\end{equation}
where \(\operatorname{deg}(\cdot)\) denotes the degree of node \(m\) for sample cell \(n\) in the graph \(\mathcal{G}^{(\text{PPI})}\).

With optional weights \(\lambda_{\mathrm{edge}},\lambda_{\mathrm{deg}}>0\), the per-sample objective and the dataset-averaged pretraining loss are
\begin{equation}
\mathcal{L}^{(n)}=\lambda_{\mathrm{edge}}\mathcal{L}_{\mathrm{edge}}^{(n)}
                 +\lambda_{\mathrm{deg}}\mathcal{L}_{\mathrm{deg}}^{(n)},
\qquad
\mathcal{L}_{\mathrm{pre}}=\frac{1}{N}\sum_{n=1}^{N}\mathcal{L}^{(n)}.
\label{eq:total-loss}
\end{equation}
This objective jointly optimizes the encoder and both decoders to reconstruct the masked PPI topology and node degrees from \(\mathcal{H}\), thereby aligning the learned embeddings with the global signaling structure.

\subsection{Model Downstream Tasks}
\label{downstream}
Ultimately, the objective is to use the pretrained foundation model, $f_{\text{pre}}(\cdot)$, that synergistically integrates the task specific incoming feature set $\mathcal{X}_{\tau}\in{\mathbb{R}^{N_{\tau}\times{M}}}$, node descriptions $\mathcal{S}$, and graph topology $\mathcal{E}$ to predict cell-specific outcomes. As to the unsupervised task, the latent embedding for the incoming feature set will be generated by
\begin{equation}
    \mathcal{H}^{(\tau)}= f_{\text{pre}}(\mathcal{X}_{\tau}, \mathcal{S}, \mathcal{E})
\end{equation}
, where $\mathcal{H}^{(\tau)}\in{\mathbb{R}^{N_{\tau}\times{M}\times{d}}}$. For supervised learning, the foundation model will predict the cell outcomes by
\begin{equation}
\hat{\mathcal{Y}_{\tau}} = f_{\text{down}}(\mathcal{H}^{(\tau)}, \mathcal{E}, \mathcal{S'})
\end{equation}
, where $\hat{\mathcal{Y}_{\tau}} \in \mathbb{R}^{N}$ represents the predicted cellular states, which depends on specific downstream tasks (e.g., cell type annotations or celluar condition (normal vs. disease)) and $f_{\text{down}}(\cdot)$ is the downstream decoder, which contains following components of encoders, 
\begin{equation}
\mathcal{H}_{\text{Cross}}^{(\tau)}=\text{ENC}^{\text{down}}_{\text{Cross}}\left(\text{ENC}^{\text{down}}_{\text{Omic}}(\mathcal{H}^{(\tau)}), \mathcal{S'}\right)
\end{equation}
\begin{equation}
\mathcal{Z}^{(\tau)} = \text{GNN}_{\text{global}}^{\text{down}}\left(\text{GNN}_{\text{in}}^{\text{down}}(\mathcal{H}_{\text{Cross}}^{(\tau)},\mathcal{E})\right)
\end{equation}
\begin{equation}
\hat{\mathcal{Y}_{\tau}} = \text{MLP}_\text{down}(\mathcal{Z}^{(\tau)})
\end{equation}
, where $\text{ENC}^{\text{down}}_{\text{Omic}}(\cdot)$ and $\text{ENC}^{\text{down}}_{\text{Cross}}(\cdot)$ represent the downstream omic encoder and downstream cross-modality encoder, $\text{GNN}_{\text{in}}^{\text{down}}(\cdot)$ and $\text{GNN}_{\text{global}}^{\text{down}}(\cdot)$ are downstream internal signaling and global signaling message propagation via GNN layers and $\text{MLP}_\text{down}(\cdot)$ is the linear classifier. To infer the cell-specific targets and signaling networks network for sample $n_{\tau}$, an affinity matrix for measuring the attention-based edge weight will be derived from $\mathcal{Z}^{(\tau)}_{n_{\tau}}$, the latent space embeddings of sample $n_{\tau}$, by
\begin{equation}
  A^{(\tau)}_{n_{\tau}} = \mathrm{ATT}\!\left(\mathcal{Z}^{(\tau)}_{n_{\tau}}\right)
\end{equation}
, where \(\mathrm{ATT}(\cdot)\) is an attention-based similarity function. To restrict the edge weights to the protein–protein interaction (PPI) topology, we use
\begin{equation}
  (W_{\mathcal{E}}^{(\tau)})_{n_{\tau}} = \text{TRAN}(A^{(\tau)}_{n_{\tau}}) \odot \text{ONEHOT}(\mathcal{E}^{(\mathrm{PPI})})
\end{equation}
, where $\text{TRAN}(\cdot)$ is the transformation function to trun the adjacency matrix to edge pairs in dimensions of $\lvert\mathcal{E}\rvert$ by 2 and \((W_{\mathcal{E}}^{(\tau)})_{n_{\tau}}\in{\mathbb{R}}^{\lvert\mathcal{E}_{\text{PPI}}\rvert\times{2}}\) are cell specific PPI edge weights for sample $n_{\tau}$. To calculate the node importance score for certain node $m$ in the sample $n_{\tau}$, we integrate the averaged bi-directional edge weights and omic values by
\begin{equation}
  \big(W_{\mathcal{V}}^{(\tau)}\big)_{n_{\tau}}^{(m)}
  =
  \sum_{i}\big[(W_{\mathcal{E}}^{(\tau)})_{mi} + (W_{\mathcal{E}}^{(\tau)})_{im}\big]
\end{equation}
, where \(\big(W_{\mathcal{V}}^{(\tau)}\big)_{n_{\tau}}^{(m)}\) are node importance score for node $m$ in sample $n_{\tau}$. Finally, a core-extraction routine will select a compact subgraph by
\begin{equation}
  \mathcal{G}^{(\tau)}_{n_{\tau}} = f_{\mathrm{core}}\!\Big(\big(W_{\mathcal{V}}^{(\tau)}\big)_{n_{\tau}},\, \big(W_{\mathcal{E}}^{(\tau)}\big)_{n_{\tau}};\, \xi,\, \epsilon\Big),
\end{equation}
, where the operator \(f_{\mathrm{core}}(\cdot)\) ranks nodes by \(\big(W_{\mathcal{V}}^{(\tau)}\big)_{n_{\tau}}\) to retain the top \(\xi\) nodes and applies a branch-pruning procedure~\cite{zhang2025mosgraphflow} to keep the top \(\epsilon\) edges in $\big(W_{\mathcal{E}}^{(\tau)}\big)_{n_{\tau}}$, yielding the core subgraph \(\mathcal{G}^{(\tau)}_{n_{\tau}}=(\mathcal{V}^{(\tau)}_{n_{\tau}},\mathcal{E}^{(\tau)}_{n_{\tau}})\).

\section*{Declarations}
\begin{itemize}
\item Funding: This research was partially supported by NIA 4R33AG078799-02, NLM 1R01LM013902-01A1, NIA R56AG065352, NIA 1R21AG078799-01A1 and NINDS 1RM1NS132962-01.
\item Competing interests: Authors has no competing interests.
\item Data availability: The OmniCellTOSG dataset is available on Hugging Face:\\
\href{https://huggingface.co/datasets/FuhaiLiAiLab/OmniCellTOSG_Dataset}{\texttt{https://huggingface.co/datasets/FuhaiLiAiLab/OmniCellTOSG\_Dataset}}. All dataset sources used in this study are documented in Data Collection (Section~\ref{data-download}). Detailed information for datasets collected from GEO and other repositories is provided in Table~\ref{data_collection}. Datasets from CELLxGENE were retrieved through the CELLxGENE Census (release version 2025-01-30) and the CELLxGENE portal\\
(\href{https://chanzuckerberg.github.io/cellxgene-census/python-api.html}{\texttt{https://chanzuckerberg.github.io/cellxgene-census/python-api.html}}, \href{https://cellxgene.cziscience.com/}{\texttt{https://cellxgene.cziscience.com/}}). The preprocessed GSE129308 dataset used in downstream evaluations and other Brain Cell Atlas data used in dataset construction were downloaded from: \href{https://www.braincellatlas.org/dataSet}{\texttt{https://www.braincellatlas.org/dataSet}}.

\item Code availability: The \texttt{CellTOSG\_Loader} package and the CellTOSG-FM foundation model code are available on GitHub:\\
\href{https://github.com/FuhaiLiAiLab/OmniCellTOSG}{\texttt{https://github.com/FuhaiLiAiLab/OmniCellTOSG}}
\item Author contribution: FL conceptualized and supervised the project. Methodology and study design were developed by HZ, TX, DC, SL, GS, NH, LS, LK, DH, CC, GL, MP, YC, PP and FL. HZ and TX led the data collection, package and model implementation and result analysis. The manuscript was written by HZ, TX, GS, NH, and FL.

\end{itemize}

\bibliography{manuscript}

\newpage
\section{Supplementary Notes}
\subsection{Data Collection} \label{data-download}

\paragraph{CZ CellxGene Database}
Data from the CELLxGENE database were obtained using the CZI Science CELLxGENE Census Python API, with the census version fixed to \texttt{2025-01-30} and the organism set to \texttt{Homo sapiens}. We retrieved matrices directly in \texttt{.h5ad} (AnnData) format, together with standardized per-cell metadata fields, to facilitate downstream harmonization and provenance tracking.

\paragraph{Brain Cell Atlas}
Human datasets were manually downloaded from the Brain Cell Atlas portal after applying the species filter \texttt{Human}. Because some processed matrices lack unique per-sample identifiers, we preserved the original project identifiers from the source datasets in the resulting \texttt{.h5ad} files. This ensures traceability across preprocessing steps and alignment with external references.

\paragraph{GEO Database and Others}
We manually retrieved datasets hosted on NCBI GEO to complement CELLxGENE and Brain Cell Atlas with broader organ and disease coverage, larger cohort sizes for our analyses. Selection prioritized (i) availability of key annotations used for harmonization and balancing (e.g., \texttt{cell\_type}, \texttt{disease}, \texttt{tissue}/\texttt{tissue\_general}, \texttt{sex}, and \texttt{suspension\_type}; \texttt{development\_stage} when available), (ii) sufficient cell counts to support robust distribution matching and downstream modeling, and (iii) feasibility of standard quality-control procedures (empty-droplet removal, doublet detection, ambient RNA mitigation) from the provided raw or study-curated matrices. We additionally incorporated two curated non-GEO resources where they provide high-quality matrices that fill specific anatomical or disease gaps in our corpus: (i) Human Cell Atlas project matrices for pancreatic tissue and retained with their original project identifiers; and (ii) the hepatitisCatlas dataset from the Broad Single Cell Portal, which supplies a curated hepatitis C cohort with consistent annotation fields. These additions are limited and purpose-driven—used only to complete organ/disease coverage and to maintain uniform preprocessing and metadata standards across the integrated collection. Accession lists and download links are provided in Table~\ref{data_collection}.

\subsection{Data Preprocessing}
\label{app:data-preproc}
To support uniform downstream analysis across heterogeneous repositories, we standardized all sources to an AnnData saved as \texttt{.h5ad}. Datasets from GEO, the Human Cell Atlas, and a hepatitis atlas arrived in diverse encodings—including Matrix Market triplets (\texttt{barcodes.tsv.gz}, \texttt{features.tsv.gz}, \texttt{matrix.mtx.gz}), compressed CSV/TXT (\texttt{.csv.gz}/\texttt{.txt.gz}), and HDF5 (\texttt{.h5})—whereas CellxGene and Brain Cell Atlas releases were largely pre-packaged as \texttt{.h5ad}. For non-\texttt{.h5ad} inputs, we reconciled barcodes, gene feature tables, and count matrices, then ingested the data with Scanpy to construct AnnData objects that preserve gene–cell mappings and available per-cell/per-gene annotations. We subsequently harmonized \texttt{obs} (cells) and \texttt{var} (genes) to a consistent schema aligned with CellxGene/Brain Cell Atlas conventions (standardized identifiers and controlled-vocabulary attributes for tissue, disease, donor, sex, age; plus gene symbols/IDs and feature types), and exported the harmonized objects to \texttt{.h5ad} for consistent I/O.

Quality checks verified internal consistency between matrix dimensions and annotation tables, the presence and datatypes of required metadata fields, and read/write integrity of the final files. The resulting curated, schema-aligned \texttt{.h5ad} datasets serve as standardized inputs for meta-cell construction and subsequent transformation into OmniCellTOSG, enabling reproducible multi-study analyses under a unified data model.

\subsection{\texttt{CellTOSG\_Loader} Package}
\subsubsection{Package Usage}\label{celltosg_loader}
For instance, row-level metadata filters (\texttt{conditions}; required) defined the cohort task retrieval scope, such as tissue- and disease-level criteria (e.g., \{\texttt{tissue\_general}: brain, \texttt{disease\_name}: Alzheimer’s Disease\}). The \texttt{task} and \texttt{label\_column} fully specifies the cohort and label semantics for downstream modeling. As to the \texttt{sample\_ratio}, this parameters will extracted this ratio of samples from whole candidate cells, since some files are pretty large and it will burst the memory storage. By using this ratio, we will sampling this ratio of cells from whole candidate cells. Following are the example code for using the package

\begin{lstlisting}[
  language=Python,
  caption={Loading the CellTOSG dataset and extracting graph-based features},
  label={lst:load_celltosg},
  basicstyle=\ttfamily\footnotesize,
  aboveskip=6pt, belowskip=4pt,
  breaklines=true, columns=fullflexible, keepspaces=true
]
from CellTOSG_Loader import CellTOSGDataLoader

# --- Build loader (uses your argparse `args`) ---
conditions = {
    "tissue_general": args.tissue_general,
    "disease_name": args.disease_name,   # or: "disease": args.disease_name
    # "suspension_type": args.suspension_type,
    # "cell_type": args.cell_type,
    # "sex": args.sex,
}

dataset = CellTOSGDataLoader(
    root=args.dataset_root,
    conditions=conditions,
    task=args.task,                          # "disease" / "sex" / "cell_type"
    label_column=args.label_column,          # "disease" / "sex" / "cell_type"
    sample_ratio=args.sample_ratio,          # mutually exclusive with sample_size
    sample_size=args.sample_size,
    shuffle=args.shuffle,
    stratified_balancing=args.stratified_balancing,
    extract_mode=args.extract_mode,          # "inference" / "train"
    train_text=args.train_text, # False -> return precomputed name/desc embeddings
    train_bio=args.train_bio,   # False -> return precomputed sequence embeddings
    correction_method=args.correction_method, # None / "combat_seq"
    output_dir=args.output_dir,
)

# --- Access tensors/arrays ---
if args.extract_mode == "inference":
    X = dataset.data                         # pandas.DataFrame (expression/features)
    y = dataset.labels                       # pandas.DataFrame
    metadata = dataset.metadata              # pandas.DataFrame (row-aligned metadata)
else:
    X = dataset.data                         # dict: {"train": X_train, "test": X_test}
    y = dataset.labels                       # dict: {"train": y_train, "test": y_test}
    metadata = dataset.metadata              # dict: {"train": meta_train, "test": meta_test}

all_edge_index = dataset.edge_index                   # full graph (COO [2, E])
internal_edge_index = dataset.internal_edge_index     # optional transcript-protein edges
ppi_edge_index = dataset.ppi_edge_index               # optional PPI edges     
x_name_emb, x_desc_emb, x_bio_emb = pre_embed_text(args, dataset, pretrain_model, device) # Prepare text and seq embeddings

\end{lstlisting}

\subsubsection{Stratified Retrieval Algorithm}
\label{sra_algo}

\FloatBarrier                
\begingroup                  
\setlength{\textfloatsep}{6pt plus 2pt minus 2pt}
\setlength{\intextsep}{6pt plus 2pt minus 2pt}
\captionsetup[algorithm]{aboveskip=0pt,belowskip=3pt}

\begin{algorithm}[!htbp]     
\caption{Stratified Retrieval Algorithm}
\begin{footnotesize}         
\begin{algorithmic}[1]
\Require Data $\mathcal{X}$; query $q$; config $\Lambda(\lambda)=(b,b_0,K,j^\star)$; tolerance $\delta$; seed $s$; upsample flag $P$
\Ensure Balanced, stratified retrieval $\mathcal{X}_{R}$

\Statex \textbf{Phase I: Query-Constrained Extraction}
\State $R(q) \gets \{\,x\in \mathcal{X}:\ \bigwedge_{(a,V_a)\in q}\ \mathcal{I}[u_a(x)\in V_a]\}$ 

\Statex \textbf{Phase II: Config-Driven Balancing}
\State $\mathrm{CA} \gets \{\,x\in R(q):\ u_b(x)\neq b_0\,\}$
\State $q_{-b} \gets \{(a,V_a)\in q:\ a\neq b\}$
\State $\mathrm{NM} \gets \{\,x\in \mathcal{X}:\ u_b(x)=b_0\ \wedge\ \bigwedge_{(a,V_a)\in q_{-b}} \mathcal{I}[u_a(x)\in V_a]\}$
\State \textbf{drop} rows in $\mathrm{CA},\mathrm{NM}$ with missing $u_{k_i}$ for any $k_i\in K$
\If{\(\lvert \mathrm{CA}\rvert \le \lvert \mathrm{NM}\rvert\)} 
  \State $\textbf{ref}\gets \mathrm{CA}$;\quad $\textbf{tgt}\gets \mathrm{NM}$
\Else
  \State $\textbf{ref}\gets \mathrm{NM}$;\quad $\textbf{tgt}\gets \mathrm{CA}$
\EndIf
\State $\mathcal{X}_{R} \gets \varnothing$ 
\ForAll{$\kappa \in \{\kappa_K(x):\ x\in \textbf{ref}\}$}
  \State \(\xi \gets \lvert\{\,x\in \mathbf{ref}:\ \kappa_K(x)=\kappa\,\}\rvert\)
  \State $\text{Match} \gets \varnothing$
  \State $\text{Used} \gets \varnothing$
  \For{$t = 0$ \textbf{to} $\delta$}
    \State $\mathrm{Cand}_t \gets \{\,x\in \textbf{tgt}:\ u_{k_i}(x)=\kappa_i\ \forall i\neq j^\star,\ d_{k^\star}(u_{k^\star}(x),\kappa_{j^\star})\le t\,\}\setminus \text{Used}$
    \State \(\eta \gets \min\{\,\xi-\lvert \mathrm{Match}\rvert,\ \lvert \mathrm{Cand}_{t}\rvert\,\}\)
    \State \textbf{sample} $\eta$ \textbf{w/o replacement} from $\mathrm{Cand}_t$ into $\mathrm{Take}$
    \State $\text{Match} \gets \text{Match} \cup \mathrm{Take}$
    \State $\text{Used} \gets \text{Used} \cup \mathrm{idx}(\mathrm{Take})$
    \If{\(\lvert \mathrm{Match}\rvert = n_\kappa\)} \Break \EndIf
  \EndFor
  \If{\(0<\lvert \mathrm{Match}\rvert<n_\kappa\ \textbf{and} \ P=\text{TRUE}\)}
    \State \(\mathrm{Match} \gets\) \((n_\kappa-\lvert \mathrm{Match}\rvert)\) draws of samples (with replacement) from \(\mathrm{Match}\) 
  \EndIf
  \If{\(\lvert \mathrm{Match}\rvert>0\)}
    \State $\mathcal{X}_{R} \gets$ $\{x\in \textbf{ref}:\ \kappa_K(x)=\kappa\} \cup \text{Match}$ 
  \EndIf
\EndFor
\State $Y_R \gets \{\ \ell_b(x)\ :\ x \in \mathcal{X}_{R}\ \}$ 
\State \Return $\mathcal{X}_{R}$, $Y_R$
\end{algorithmic}
\end{footnotesize}
\end{algorithm}
\endgroup

\subsubsection{Train-test Datasets Split}
In order to avoid subject-level data leakage, we performed a donor-level split in which all samples from the same donor were assigned exclusively to either the training set or the test set. A unique donor identifier was constructed by combining the study identifier with the donor identifier, ensuring that donors remain distinct even when the same donor label may appear in different studies. For each cohort, we selected a set of test donors to achieve an approximately fixed test set size at the sample level, while imposing an explicit upper bound to prevent the test set from being dominated by donors contributing unusually large numbers of samples. Donors were processed in a randomized order with a preference for smaller donors to better match the intended test proportion.

\subsection{External Dataset}
\label{external-ad-processing}
We used a Brain Cell Atlas preprocessed GSE129308 AnnData (H5AD) file as an external Alzheimer's disease (AD) cohort and applied a unified preprocessing pipeline across disease, sex, and cell type prediction tasks. Cells lacking donor identifiers or cell type annotations were excluded, and disease and sex labels were curated by removing unknown or unannotated entries and normalizing them into binary targets (control versus AD for disease, female versus male for sex). Train and test sets were defined at the donor level to eliminate data leakage, ensuring that all cells from a given donor were assigned exclusively to a single split. For binary tasks (disease and sex), donors were selected based on cell counts to match the desired test fraction while approximately preserving label prevalence under a strict cap on test set size, whereas for cell type prediction donors were partitioned solely according to cell counts. Consistent label coverage across splits was maintained by retaining only the intersection of cell types shared between training and test sets. Class imbalance was addressed through controlled downsampling to obtain approximately balanced class distributions in both partitions. The processed H5AD files were subsequently converted into fixed width NumPy expression matrices by mapping gene identifiers (gene symbols or Ensembl IDs) to a predefined BioMedGraphica feature index space, with unmapped features zero padded. Expression values were library size normalized to a fixed target sum and log1p transformed. Labels were exported as encoded NumPy arrays together with mapping tables to ensure reproducibility. The resulting feature matrices and label vectors are directly compatible with the input requirements of CellTOSG-FM for downstream modeling.

\newpage

\section{Supplementary Tables}
\setcounter{table}{0}
\makeatletter 
\renewcommand{\thetable}{S\@arabic\c@table}
\makeatother

\newcolumntype{P}[1]{>{\RaggedRight\arraybackslash}p{#1}}
\newcolumntype{Q}[1]{>{\centering\arraybackslash}p{#1}}
\newcolumntype{X}{>{\RaggedRight\arraybackslash}p{\linewidth}}

\begin{table}[h!]
    \centering
    \caption{Detailed information on datasets collected from GEO and other sources}
    \label{data_collection}
    \resizebox{\linewidth}{!}{%
    \renewcommand{\arraystretch}{1.0}
    \begin{tabular}{
        P{2.5cm}
        P{4.0cm}
        P{8.0cm}
        P{1.5cm}
        Q{2.0cm}
        P{6.0cm}
    }
    \toprule
    \textbf{Platform} & \textbf{Dataset ID} & \textbf{Disease(s)} &
    \textbf{Tissue General}  & \textbf{Cell Count} & \textbf{Links} \\
    \midrule
    GEO & GSE183852 &
    Dilated cardiomyopathy & Heart & \num{49723} &
    \url{https://www.ncbi.nlm.nih.gov/geo/query/acc.cgi?acc=GSE183852} \\
    GEO & GSE125449 &
    Hepatocellular carcinoma; Intrahepatic Cholangiocarcinoma & Liver & \num{9946} &
    \url{https://www.ncbi.nlm.nih.gov/geo/query/acc.cgi?acc=GSE125449} \\
    Single Cell Portal & hepatitisCatlas &
    Hepatitis C & Liver, Blood & \num{8350} &
    \url{https://singlecell.broadinstitute.org/single_cell/study/SCP2407/single-cell-atlas-of-the-liver-myeloid-compartment-before-and-after-cure-of-chronic-viral-hepatitis\#study-download} \\
    Human Cell Atlas & PancreasTopographiesTosti10x &
    Carcinosarcoma; Chronic Pancreatitis; Pancreatic Ductal Adenocarcinoma; Pancreatic Neuroendocrine Neoplasm &
    Pancreas & \num{125757} &
    \url{https://explore.data.humancellatlas.org/projects/b3938158-4e8d-4fdb-9e13-9e94270dde16/project-matrices} \\
    \bottomrule
    \end{tabular}%
    }
\end{table}

\newpage
\begin{table}[htbp]
\centering
\tiny
\setlength{\tabcolsep}{4pt}
\renewcommand{\arraystretch}{1.25}
\caption{Description of the final attribute set $\mathcal{A}$.}
\label{tab:attributes}
\begin{tabular}{@{}l p{3.6cm} l c p{2.8cm}@{}}
\toprule
\textbf{Attribute} & \textbf{Meaning} & \textbf{Type} & \textbf{Req.} & \textbf{Example} \\
\midrule
source & Origin of the record or data repository/provider. & string & No & CellxGene, BrainCellAtlas \\

dataset\_id & Unique dataset identifier within/among sources. & string & Yes & GSE144744, EGAS00001004107 \\

suspension\_type & sc/snRNA-seq. & string & Yes & cell, nucleus \\

tissue\_general (coarse-grained) & Tissue/organ category. & string & Yes* & brain, liver \\

tissue (fine-grained) & Specific tissue/region. & string & No & prefrontal cortex, heart right ventricle \\

matrix\_file\_path & File relative path to the expression matrix file. & string & Yes & /expression\_matrix/ braincellatlas/brain\_part\_0.npy \\

matrix\_row\_idx & Row index in the matrix for this entity/sample. & integer & Yes & 2025 \\

donor\_id & Donor ID. & string & Yes & Donor26 \\

CMT\_id & Cell type ID mapped to Cell Ontology. & string & No & CMT0151 \\

CMT\_name & Cell type name mapped to Cell Ontology. & string & No & microglial cell \\

disease\_BMG\_name & Standardized disease label mapped to BioMedGraphica terminology. & string & Yes* & Alzheimer’s disease \\

disease\_BMG\_id & Standardized disease ID mapped to BioMedGraphica ID. & string & Yes* & BMGC\_DS00092 \\

development\_stage\_category & Broad development stage category. & enum & No & embryo, fetal, adult \\

sex\_normalized & Normalized sex label after harmonization. & enum & Yes & male, female, unknown \\
\bottomrule
\end{tabular}

\vspace{1pt}
\tiny
\textbf{Notes:} “Req.” = required field. Yes = mandatory; No = optional; Yes* = required in most cases (e.g., for healthy/control, age \& sex cohorts).
\end{table}

\newpage
{\tiny
\setlength{\LTpre}{0pt}
\setlength{\LTpost}{0pt}
\setlength{\tabcolsep}{4pt}
\renewcommand{\arraystretch}{1.15}

\begin{longtable}{
  >{\raggedright\arraybackslash}p{3.5cm}
  >{\raggedright\arraybackslash}p{4.5cm}
  >{\centering\arraybackslash}p{1.5cm}
  >{\centering\arraybackslash}p{1.1cm}
}
\caption{OmniCellTOSG Dataset Overview and Detailed Statistics}
\label{tab:data_overview} \\
\toprule
\textbf{Diseases} & \textbf{Organ/Tissue Types} & \textbf{\# of Original Cells} & \textbf{\# of Meta Cells} \\
\midrule
\endfirsthead
\toprule
\textbf{Diseases} & \textbf{Organ/Tissue Types} & \textbf{\# of Original Cells} & \textbf{\# of Meta Cells} \\
\midrule
\endhead
Normal & Multiple Tissue* & 56,283,477 & 280,978 \\
Covid-19 & Blood, Brain, Digestive System, Lung, Nose, Respiratory System, Saliva & 5,130,024 & 25,621 \\
Parkinson Disease & Brain & 1,736,202 & 8,675 \\
Alzheimer's Disease & Brain & 1,665,182 & 8,316 \\
Glioblastoma & Brain & 1,236,376 & 6,171 \\
Dementia & Brain & 1,052,021 & 5,259 \\
Malignant Ovarian Serous Tumor & Abdomen, Colon, Fallopian Tube, Intestine, Large Intestine, Liver, Lymph Node, Musculature, Omentum, Ovary, Paracolic Gutter, Urinary Bladder, Uterus & 927,205 & 4,629 \\
Lung Adenocarcinoma & Adrenal Gland, Brain, Liver, Lung, Lymph Node, Pleural Fluid & 862,494 & 4,304 \\
Systemic Lupus Erythematosus & Blood & 777,258 & 3,886 \\
Crohn Disease & Colon, Small Intestine & 572,174 & 2,844 \\
Breast Cancer & Axilla, Brain, Breast, Chest Wall, Liver, Lung, Neck, Skeletal System, Skin Of Body & 553,514 & 2,760 \\
Dilated Cardiomyopathy & Heart & 519,709 & 2,596 \\
Multiple Sclerosis & Blood, Brain & 442,149 & 2,210 \\
Chronic Kidney Disease & Kidney & 370,831 & 1,851 \\
Amyotrophic Lateral Sclerosis & Brain & 321,851 & 1,608 \\
Atrial Fibrillation & Heart & 273,963 & 1,369 \\
Atherosclerosis & Heart, Vasculature & 259,721 & 1,297 \\
Frontotemporal Dementia & Brain & 252,352 & 1,260 \\
Temporal Lobe Epilepsy & Brain & 246,167 & 1,230 \\
Squamous Cell Lung Carcinoma & Lung & 243,071 & 1,214 \\
Pulmonary Fibrosis & Lung & 218,932 & 1,094 \\
Nonpapillary Renal Cell Carcinoma & Adrenal Gland, Kidney, Vasculature & 207,496 & 1,035 \\
Acute Kidney Failure & Kidney & 189,295 & 944 \\
Clear Cell Renal Carcinoma & Blood, Kidney, Lymph Node & 187,792 & 938 \\
B-cell Acute Lymphoblastic Leukemia & Blood & 183,023 & 913 \\
Triple-negative Breast Carcinoma & Breast, Exocrine Gland, Pleural Fluid & 179,724 & 897 \\
Gliomas & Brain & 169,399 & 845 \\
Chronic Obstructive Pulmonary Disease & Lung, Respiratory System & 164,361 & 820 \\
Myocardial Infarction & Brain, Heart & 152,055 & 760 \\
Epilepsy & Brain & 134,460 & 671 \\
Follicular Lymphoma & Lymph Node & 122,702 & 613 \\
Non-small Cell Lung Carcinoma & Lung & 120,796 & 603 \\
Pancreatic Ductal Adenocarcinoma & Pancreas & 118,676 & 592 \\
Progressive Supranuclear Palsy & Brain & 117,482 & 585 \\
Basal Cell Carcinoma & Skin Of Body & 115,456 & 573 \\
Unknown & Blood & 109,488 & 547 \\
Primary Sclerosing Cholangitis & Liver & 104,667 & 522 \\
Arrhythmogenic Right Ventricular Cardiomyopathy & Heart & 104,496 & 522 \\
Unclassified & Brain & 99,009 & 494 \\
Pick Disease & Brain & 98,043 & 489 \\
Post-covid-19 Disorder & Blood & 97,224 & 486 \\
Respiratory System Disorder & Blood & 94,987 & 474 \\
Digestive System Disorder & Small Intestine & 89,849 & 449 \\
Estrogen-receptor Positive Breast Cancer & Breast & 87,648 & 438 \\
Invasive Ductal Breast Carcinoma & Breast, Exocrine Gland & 86,873 & 433 \\
Juvenile Dermatomyositis & Blood & 82,666 & 413 \\
Small Cell Lung Carcinoma & Adrenal Gland, Axilla, Brain, Liver, Lung, Lymph Node, Pleural Fluid & 79,040 & 390 \\
Amyotrophic Lateral Sclerosis 26 With Or Without Frontotemporal Dementia & Brain & 73,797 & 368 \\
Interstitial Lung Disease & Lung & 68,456 & 342 \\
Benign Prostatic Hyperplasia & Prostate Gland & 66,181 & 330 \\
Lewy Body Dementia & Brain & 65,789 & 327 \\
Common Variable Immunodeficiency & Blood & 64,081 & 319 \\
B-cell Non-hodgkin Lymphoma & Bone Marrow & 59,746 & 298 \\
Blastoma & Liver & 57,445 & 287 \\
Opiate Dependence & Brain & 54,399 & 271 \\
Autism Spectrum Disorder & Brain & 52,003 & 260 \\
Oropharynx Squamous Cell Carcinoma & Digestive System & 50,000 & 250 \\
Rheumatoid Arthritis & Blood & 48,637 & 243 \\
Clonal Hematopoiesis & Blood & 47,354 & 236 \\
Luminal B Breast Carcinoma & Pleural Fluid & 46,128 & 229 \\
Sjogren Syndrome & Exocrine Gland & 45,231 & 226 \\
Major Depressive Disorder & Brain & 41,944 & 209 \\
Primary Biliary Cholangitis & Liver & 39,994 & 199 \\
Periodontitis & Mucosa & 38,520 & 192 \\
Influenza & Blood, Brain & 34,549 & 171 \\
Pilocytic Astrocytoma & Brain & 34,291 & 171 \\
Crohn Ileitis & Small Intestine & 32,458 & 162 \\
Pneumonia & Lung & 31,923 & 159 \\
Pulmonary Emphysema & Lung & 31,792 & 158 \\
Diffuse Large B-cell Lymphoma & Lymph Node, Respiratory System, Small Intestine & 31,131 & 153 \\
Chronic Rhinitis & Nose & 29,137 & 145 \\
Listeriosis & Placenta & 28,237 & 141 \\
Oral Cavity Squamous Cell Carcinoma & Digestive System & 28,186 & 140 \\
Toxoplasmosis & Placenta & 28,098 & 140 \\
Plasmodium Malariae Malaria & Placenta & 27,958 & 139 \\
Acute Myeloid Leukemia & Bone Marrow & 27,852 & 139 \\
Gastric Intestinal Metaplasia & Stomach & 27,462 & 136 \\
Gastritis & Stomach & 26,639 & 133 \\
Hiv Infectious Disease & Blood & 24,548 & 122 \\
Type 1 Diabetes Mellitus & Pancreas & 22,400 & 112 \\
Lung Large Cell Carcinoma & Lung & 21,167 & 105 \\
Tubular Adenoma & Colon & 20,442 & 99 \\
Luminal A Breast Carcinoma & Pleural Fluid & 20,403 & 101 \\
Type 2 Diabetes Mellitus & Kidney, Vasculature & 19,262 & 95 \\
Leukoencephalopathy, Diffuse Hereditary, With Spheroids 1 & Brain & 19,164 & 93 \\
Idiopathic Parkinson's Disease & Brain & 19,002 & 95 \\
Hydrosalpinx & Fallopian Tube & 17,798 & 88 \\
Cystic Fibrosis & Lung & 17,590 & 87 \\
Trisomy 18 & Brain & 16,900 & 84 \\
Down Syndrome & Bone Marrow & 16,743 & 83 \\
Invasive Lobular Breast Carcinoma & Breast, Exocrine Gland & 16,507 & 81 \\
Her2 Positive Breast Carcinoma & Breast & 16,017 & 80 \\
Squamous Cell Carcinoma & Skin Of Body & 15,328 & 75 \\
Secondary Progressive Multiple Sclerosis & Brain & 14,469 & 72 \\
Cell Stress & Brain & 13,165 & 65 \\
Melanoma & Skin Of Body & 13,141 & 65 \\
Kidney Oncocytoma & Kidney & 12,610 & 63 \\
Lymphangioleiomyomatosis & Lung & 12,374 & 61 \\
Relapsing-remitting Multiple Sclerosis & Brain & 12,201 & 61 \\
Non-compaction Cardiomyopathy & Heart & 11,632 & 57 \\
Adenocarcinoma & Colon, Large Intestine, Small Intestine & 11,483 & 55 \\
Intrahepatic Cholangiocarcinoma & Liver & 11,466 & 57 \\
Barrett Esophagus & Esophagus & 10,952 & 54 \\
Metastatic Melanoma & Brain & 10,895 & 54 \\
Colon Sessile Serrated Adenoma/polyp & Colon, Intestine & 10,893 & 53 \\
Pleomorphic Carcinoma & Lung & 10,765 & 53 \\
Hypersensitivity Pneumonitis & Lung & 10,379 & 51 \\
Aspiration Pneumonia & Brain & 10,204 & 51 \\
Non-specific Interstitial Pneumonia & Lung & 8,597 & 42 \\
Acute Myocardial Infarction & Brain & 8,033 & 40 \\
Carcinosarcoma & Pancreas & 7,967 & 39 \\
Hepatitis C Virus Infection & Liver & 7,607 & 38 \\
Breast Carcinoma & Breast & 7,373 & 36 \\
Hepatocellular Carcinoma & Liver & 7,165 & 35 \\
Primary Cutaneous Diffuse Large B-cell Lymphoma, Leg Type & Skin Of Body & 7,097 & 35 \\
Tubulovillous Adenoma & Colon, Intestine & 6,793 & 32 \\
Gingivitis & Mucosa & 6,587 & 32 \\
Injury & Skin Of Body & 5,983 & 29 \\
Mixed Gliomas & Brain & 5,979 & 29 \\
Anencephaly & Lung & 5,499 & 27 \\
Pancreatic Neuroendocrine Neoplasm & Pancreas & 5,233 & 26 \\
Congenital Heart Disease & Brain & 5,046 & 25 \\
Pulmonary Sarcoidosis & Lung & 4,886 & 24 \\
Wilms Tumor & Kidney & 4,636 & 23 \\
Heart Failure & Brain & 4,594 & 22 \\
Acute Promyelocytic Leukemia & Bone Marrow & 3,734 & 18 \\
Pulpitis & Skeletal System & 3,655 & 18 \\
Respiratory Failure & Digestive System & 3,335 & 16 \\
Anaplastic Astrocytoma & Brain & 3,097 & 15 \\
Macular Degeneration & Eye & 3,011 & 15 \\
Tongue Cancer & Brain & 2,992 & 14 \\
Mild Cognitive Impairment & Brain & 2,851 & 14 \\
Chronic Pancreatitis & Pancreas & 2,666 & 13 \\
Neuroendocrine Carcinoma & Small Intestine & 2,623 & 13 \\
Hyperplastic Polyp & Colon & 2,616 & 13 \\
Chromophobe Renal Cell Carcinoma & Kidney & 2,576 & 12 \\
Long Covid-19 & Digestive System & 2,306 & 11 \\
Colorectal Cancer & Colon & 2,199 & 10 \\
Malignant Pancreatic Neoplasm & Brain & 2,148 & 10 \\
Enamel Caries & Skeletal System & 2,015 & 10 \\
Heart Disorder & Brain & 1,957 & 9 \\
Cataract & Eye & 1,810 & 9 \\
\midrule
\multicolumn{2}{>{\raggedright\arraybackslash}p{8cm}}{\textbf{Total}} & 79,195,364 & 395,317 \\
\bottomrule
\end{longtable}

\noindent
\raggedright
\fontsize{3}{4}\selectfont
\\
\textit{* Multiple Tissue (Normal):} Adipose Tissue, Adrenal Gland, Bladder Organ, Blood, Bone Marrow, Brain, Breast, Central Nervous System, Colon, Cortex, Digestive System, Embryo, Endocrine Gland, Esophagogastric Junction, Esophagus, Exocrine Gland, Eye, Fallopian Tube, Forelimb, Gallbladder, Head, Heart, Hindlimb, Immune System, Intestine, Kidney, Lamina Propria, Large Intestine, Liver, Lung, Lymph Node, Milk, Mucosa, Musculature, Nose, Omentum, Ovary, Pancreas, Placenta, Pleura, Prostate Gland, Respiratory System, Scalp, Sensory System, Skeletal System, Skin Of Body, Small Intestine, Spinal Cord, Spleen, Stomach, Tendon Of Semitendinosus, Testis, Tongue, Ureter, Urinary Bladder, Uterus, Vasculature, Yolk Sac

}

\newpage
\begin{table}[th]
\caption{Model performance across diseases on the cell-type annotation task}
\vspace{-0.00in}
\label{tab:celltype-performance}
\begin{center}
\resizebox{\textwidth}{!}{%
\begin{tabular}{lc*{8}{c}}

\toprule
\textbf{Disease} & \textbf{Metric} & \textbf{GCN} & \textbf{GAT} & \textbf{UniMP} & \textbf{DNN} & \textbf{scCELLO} & \textbf{scFoundation} & \textbf{scGPT} & \textbf{CellTOSG-FM} \\
\midrule
\multirow{2}{*}{\textbf{AD}} & Accuracy & $ 0.2353 \scriptstyle \pm 0.1544 $ & $ 0.0537 \scriptstyle \pm 0.0107 $ & $ 0.4404 \scriptstyle \pm 0.0279 $ & $ 0.7145 \scriptstyle \pm 0.0334 $ & $ 0.7788 \scriptstyle \pm 0.0164 $ & $ 0.7719 \scriptstyle \pm 0.0264 $ & $ 0.7944 \scriptstyle \pm 0.0108 $ & \boldmath{$ 0.8792 \scriptstyle \pm 0.0164 $} \\
 & F1 & $ 0.0394 \scriptstyle \pm 0.0236 $ & $ 0.0113 \scriptstyle \pm 0.0021 $ & $ 0.0747 \scriptstyle \pm 0.0089 $ & $ 0.3108 \scriptstyle \pm 0.0286 $ & $ 0.5669 \scriptstyle \pm 0.0258 $ & $ 0.4486 \scriptstyle \pm 0.0631 $ & $ 0.4234 \scriptstyle \pm 0.0260 $ & \boldmath{$ 0.6997 \scriptstyle \pm 0.0838 $} \\
\midrule
\multirow{2}{*}{\textbf{LUAD}} & Accuracy & $ 0.0733 \scriptstyle \pm 0.0085 $ & $ 0.0583 \scriptstyle \pm 0.0047 $ & $ 0.0700 \scriptstyle \pm 0.0187 $ & $ 0.4333 \scriptstyle \pm 0.0306 $ & $ 0.2067 \scriptstyle \pm 0.0368 $ & $ 0.5133 \scriptstyle \pm 0.0266 $ & $ 0.2467 \scriptstyle \pm 0.0978 $ & \boldmath{$ 0.5767 \scriptstyle \pm 0.0628 $} \\
 & F1 & $ 0.0175 \scriptstyle \pm 0.0020 $ & $ 0.0155 \scriptstyle \pm 0.0027 $ & $ 0.0112 \scriptstyle \pm 0.0051 $ & $ 0.3615 \scriptstyle \pm 0.0366 $ & $ 0.1649 \scriptstyle \pm 0.0355 $ & $ 0.4489 \scriptstyle \pm 0.0402 $ & $ 0.2057 \scriptstyle \pm 0.1082 $ & \boldmath{$ 0.5247 \scriptstyle \pm 0.0672 $} \\
\midrule
\multirow{2}{*}{\textbf{AF}} & Accuracy & $ 0.1810 \scriptstyle \pm 0.0488 $ & $ 0.1732 \scriptstyle \pm 0.0203 $ & $ 0.2500 \scriptstyle \pm 0.0000 $ & $ 0.8242 \scriptstyle \pm 0.0096 $ & $ 0.8737 \scriptstyle \pm 0.0103 $ & $ 0.8802 \scriptstyle \pm 0.0488 $ & $ 0.8112 \scriptstyle \pm 0.0112 $ & \boldmath{$ 0.9792 \scriptstyle \pm 0.0018 $} \\
 & F1 & $ 0.0405 \scriptstyle \pm 0.0075 $ & $ 0.0368 \scriptstyle \pm 0.0037 $ & $ 0.0500 \scriptstyle \pm 0.0000 $ & $ 0.6079 \scriptstyle \pm 0.0367 $ & $ 0.7431 \scriptstyle \pm 0.0173 $ & $ 0.8391 \scriptstyle \pm 0.0942 $ & $ 0.7078 \scriptstyle \pm 0.0220 $ & \boldmath{$ 0.9744 \scriptstyle \pm 0.0063 $} \\
\midrule
\multirow{2}{*}{\textbf{SLE}} & Accuracy & $ 0.1383 \scriptstyle \pm 0.0230 $ & $ 0.2482 \scriptstyle \pm 0.0591 $ & $ 0.2801 \scriptstyle \pm 0.0779 $ & $ 0.7305 \scriptstyle \pm 0.0887 $ & $ 0.5674 \scriptstyle \pm 0.0509 $ & $ 0.8475 \scriptstyle \pm 0.0558 $ & $ 0.8227 \scriptstyle \pm 0.1248 $ & \boldmath{$ 0.8972 \scriptstyle \pm 0.0929 $} \\
 & F1 & $ 0.0342 \scriptstyle \pm 0.0042 $ & $ 0.0905 \scriptstyle \pm 0.0278 $ & $ 0.0978 \scriptstyle \pm 0.0435 $ & $ 0.6678 \scriptstyle \pm 0.1201 $ & $ 0.3990 \scriptstyle \pm 0.0629 $ & $ 0.8449 \scriptstyle \pm 0.0615 $ & $ 0.8111 \scriptstyle \pm 0.1415 $ & \boldmath{$ 0.8617 \scriptstyle \pm 0.1463 $} \\
\addlinespace[2pt]
\bottomrule

\end{tabular}%
}
\end{center}
\vspace{-0.05in}
\end{table}

\newpage

\begin{table}[h]
\caption{Performance of models on cell conditions (disease vs. normal) across diseases}
\vspace{-0.00in}
\label{tab:disease-performance}
\begin{center}
\resizebox{0.8\textwidth}{!}{%
\begin{tabular}{lc*{5}{c}}

\toprule
\textbf{Disease} & \textbf{Metric} & \textbf{GCN} & \textbf{GAT} & \textbf{UniMP} & \textbf{DNN} & \textbf{CellTOSG-FM} \\
\midrule
\multirow{2}{*}{\textbf{AD}} & Accuracy & $ 0.5045 \scriptstyle \pm 0.0064 $ & $ 0.4864 \scriptstyle \pm 0.0099 $ & $ 0.5407 \scriptstyle \pm 0.0289 $ & $ 0.6957 \scriptstyle \pm 0.0330 $ & \boldmath{$ 0.7913 \scriptstyle \pm 0.0147 $} \\
 & F1 & $ 0.3928 \scriptstyle \pm 0.0841 $ & $ 0.3821 \scriptstyle \pm 0.0635 $ & $ 0.4697 \scriptstyle \pm 0.0968 $ & $ 0.6909 \scriptstyle \pm 0.0366 $ & \boldmath{$ 0.7894 \scriptstyle \pm 0.0151 $} \\
\midrule
\multirow{2}{*}{\textbf{LUAD}} & Accuracy & $ 0.5774 \scriptstyle \pm 0.0539 $ & $ 0.6730 \scriptstyle \pm 0.0346 $ & $ 0.7111 \scriptstyle \pm 0.0299 $ & $ 0.7972 \scriptstyle \pm 0.0132 $ & \boldmath{$ 0.8400 \scriptstyle \pm 0.0152 $} \\
 & F1 & $ 0.4952 \scriptstyle \pm 0.1167 $ & $ 0.6472 \scriptstyle \pm 0.0442 $ & $ 0.7019 \scriptstyle \pm 0.0374 $ & $ 0.7902 \scriptstyle \pm 0.0134 $ & \boldmath{$ 0.8381 \scriptstyle \pm 0.0150 $} \\
\midrule
\multirow{2}{*}{\textbf{AF}} & Accuracy & $ 0.5231 \scriptstyle \pm 0.0327 $ & $ 0.5046 \scriptstyle \pm 0.0065 $ & $ 0.4884 \scriptstyle \pm 0.0657 $ & $ 0.6944 \scriptstyle \pm 0.0354 $ & \boldmath{$ 0.9653 \scriptstyle \pm 0.0204 $} \\
 & F1 & $ 0.4116 \scriptstyle \pm 0.1107 $ & $ 0.3470 \scriptstyle \pm 0.0193 $ & $ 0.3931 \scriptstyle \pm 0.0595 $ & $ 0.6826 \scriptstyle \pm 0.0366 $ & \boldmath{$ 0.9653 \scriptstyle \pm 0.0204 $} \\
\midrule
\multirow{2}{*}{\textbf{SLE}} & Accuracy & $ 0.5556 \scriptstyle \pm 0.0109 $ & $ 0.5983 \scriptstyle \pm 0.1013 $ & $ 0.5940 \scriptstyle \pm 0.0757 $ & $ 0.8397 \scriptstyle \pm 0.1180 $ & \boldmath{$ 0.9979 \scriptstyle \pm 0.0030 $} \\
 & F1 & $ 0.4199 \scriptstyle \pm 0.0877 $ & $ 0.5250 \scriptstyle \pm 0.1531 $ & $ 0.5498 \scriptstyle \pm 0.0712 $ & $ 0.8265 \scriptstyle \pm 0.1339 $ & \boldmath{$ 0.9978 \scriptstyle \pm 0.0031 $} \\
\addlinespace[2pt]
\bottomrule

\end{tabular}%
}
\end{center}
\vspace{-0.05in}
\end{table}

\newpage

\begin{table}[h]
\caption{Performance of models on cell sex classifications}
\vspace{-0.00in}
\label{tab:sex-performance}
\begin{center}
\resizebox{0.8\textwidth}{!}{%
\begin{tabular}{lc*{5}{c}}

\toprule
\textbf{Disease} & \textbf{Metric} & \textbf{GCN} & \textbf{GAT} & \textbf{UniMP} & \textbf{DNN} & \textbf{CellTOSG-FM} \\
\midrule
\multirow{2}{*}{\textbf{AD}} & Accuracy & $ 0.5098 \scriptstyle \pm 0.0100 $ & $ 0.5020 \scriptstyle \pm 0.0028 $ & $ 0.5059 \scriptstyle \pm 0.0144 $ & $ 0.6588 \scriptstyle \pm 0.0546 $ & \boldmath{$ 0.7882 \scriptstyle \pm 0.0144 $} \\
 & F1 & $ 0.3730 \scriptstyle \pm 0.0411 $ & $ 0.3409 \scriptstyle \pm 0.0107 $ & $ 0.4359 \scriptstyle \pm 0.0459 $ & $ 0.6456 \scriptstyle \pm 0.0624 $ & \boldmath{$ 0.7864 \scriptstyle \pm 0.0147 $} \\
\midrule
\multirow{2}{*}{\textbf{AF}} & Accuracy & $ 0.5580 \scriptstyle \pm 0.0256 $ & $ 0.5072 \scriptstyle \pm 0.0489 $ & $ 0.4529 \scriptstyle \pm 0.0505 $ & $ 0.4964 \scriptstyle \pm 0.0051 $ & \boldmath{$ 0.9022 \scriptstyle \pm 0.0307 $} \\
 & F1 & $ 0.4795 \scriptstyle \pm 0.0704 $ & $ 0.4401 \scriptstyle \pm 0.0632 $ & $ 0.3998 \scriptstyle \pm 0.0540 $ & $ 0.3317 \scriptstyle \pm 0.0023 $ & \boldmath{$ 0.9019 \scriptstyle \pm 0.0307 $} \\
\addlinespace[2pt]
\bottomrule

\end{tabular}%
}
\end{center}
\vspace{-0.05in}
\end{table}

\newpage

\begin{table}[h]
\caption{Model performances on cell sex classfication across age groups in AF}
\vspace{-0.00in}
\label{tab:af-sex-age}
\begin{center}
\resizebox{0.85\textwidth}{!}{%
\begin{tabular}{lc*{10}{c}}

\toprule
\textbf{Age Group} & \textbf{Metric}  & \textbf{GCN} & \textbf{GAT} & \textbf{UniMP} & \textbf{DNN} & \textbf{CellTOSG-FM} \\
\midrule
\multirow{2}{*}{\textbf{Aged (\text{$\geq$} 65yrs)}} & Accuracy & $0.5897 \scriptstyle \pm 0.0363$ & $0.5385 \scriptstyle \pm 0.1088$ & $0.4359 \scriptstyle \pm 0.0725$ & $0.5385 \scriptstyle \pm 0.1088$ & \boldmath{$0.6410 \scriptstyle \pm 0.0725$} \\
 & F1 & $0.6025 \scriptstyle \pm 0.1143$ & $0.4996 \scriptstyle \pm 0.3534$ & $0.3762 \scriptstyle \pm 0.2882$ & $0.5079 \scriptstyle \pm 0.3592$ & \boldmath{$0.7593 \scriptstyle \pm 0.0429$} \\
\midrule
\multirow{2}{*}{\textbf{80 and over}} & Accuracy & $0.5527 \scriptstyle \pm 0.0332$ & $0.5021 \scriptstyle \pm 0.0418$ & $0.4557 \scriptstyle \pm 0.0474$ & $0.4895 \scriptstyle \pm 0.0215$ & \boldmath{$0.9451 \scriptstyle \pm 0.0239$} \\
 & F1 & $0.4533 \scriptstyle \pm 0.1468$ & $0.4603 \scriptstyle \pm 0.2409$ & $0.3668 \scriptstyle \pm 0.2227$ & $0.4292 \scriptstyle \pm 0.3035$ & \boldmath{$0.9436 \scriptstyle \pm 0.0261$} \\
\bottomrule

\end{tabular}%
}
\end{center}
\vspace{-0.05in}
\end{table}

\newpage

\begin{table}[h]
\caption{Model performances on sex classfication across cell types in AF}
\vspace{-0.00in}
\label{tab:af-sex-celltype}
\begin{center}
\resizebox{0.95\textwidth}{!}{%
\begin{tabular}{lc*{12}{c}}

\toprule
\textbf{Cell Type} & \textbf{Metric}  & \textbf{GCN} & \textbf{GAT} & \textbf{UniMP} & \textbf{DNN} & \textbf{CellTOSG-FM} \\
\midrule
\multirow{2}{*}{Cardiac muscle cell} & Accuracy & $0.6333 \scriptstyle \pm 0.0943$ & $0.4333 \scriptstyle \pm 0.0624$ & $0.3833 \scriptstyle \pm 0.0624$ & $0.5000 \scriptstyle \pm 0.0000$ & \boldmath{$0.9000 \scriptstyle \pm 0.0707$} \\
 & F1 & $0.6210 \scriptstyle \pm 0.0390$ & $0.3028 \scriptstyle \pm 0.2421$ & $0.2451 \scriptstyle \pm 0.2525$ & $0.4444 \scriptstyle \pm 0.3143$ & \boldmath{$0.9076 \scriptstyle \pm 0.0633$} \\
\midrule
\multirow{2}{*}{Cardiac blood vessel endothelial cell} & Accuracy & $0.5000 \scriptstyle \pm 0.0000$ & $0.4444 \scriptstyle \pm 0.0786$ & $0.4630 \scriptstyle \pm 0.0262$ & $0.5000 \scriptstyle \pm 0.0000$ & \boldmath{$1.0000 \scriptstyle \pm 0.0000$} \\
 & F1 & $0.2222 \scriptstyle \pm 0.3143$ & $0.4921 \scriptstyle \pm 0.2469$ & $0.4274 \scriptstyle \pm 0.3029$ & $0.4444 \scriptstyle \pm 0.3143$ & \boldmath{$1.0000 \scriptstyle \pm 0.0000$} \\
\midrule
\multirow{2}{*}{Fibroblast} & Accuracy & $0.4815 \scriptstyle \pm 0.0262$ & $0.5185 \scriptstyle \pm 0.0944$ & $0.4259 \scriptstyle \pm 0.0944$ & $0.5000 \scriptstyle \pm 0.0000$ & \boldmath{$0.8519 \scriptstyle \pm 0.0262$} \\
 & F1 & $0.2222 \scriptstyle \pm 0.3143$ & $0.5221 \scriptstyle \pm 0.2606$ & $0.3932 \scriptstyle \pm 0.2239$ & $0.4444 \scriptstyle \pm 0.3143$ & \boldmath{$0.8677 \scriptstyle \pm 0.0150$} \\
\midrule
\multirow{2}{*}{Adipocyte} & Accuracy & $0.6333 \scriptstyle \pm 0.0943$ & $0.6000 \scriptstyle \pm 0.0816$ & $0.5000 \scriptstyle \pm 0.0000$ & $0.5000 \scriptstyle \pm 0.0000$ & \boldmath{$0.9000 \scriptstyle \pm 0.0000$} \\
 & F1 & $0.7350 \scriptstyle \pm 0.0483$ & $0.4091 \scriptstyle \pm 0.3038$ & $0.1481 \scriptstyle \pm 0.2095$ & $0.4444 \scriptstyle \pm 0.3143$ & \boldmath{$0.9091 \scriptstyle \pm 0.0000$} \\
\midrule
\multirow{2}{*}{Mesothelial cell} & Accuracy & $0.7083 \scriptstyle \pm 0.1559$ & $0.6667 \scriptstyle \pm 0.2357$ & $0.6250 \scriptstyle \pm 0.1021$ & $0.5000 \scriptstyle \pm 0.0000$ & \boldmath{$1.0000 \scriptstyle \pm 0.0000$} \\
 & F1 & $0.7302 \scriptstyle \pm 0.0898$ & $0.5556 \scriptstyle \pm 0.4157$ & $0.3556 \scriptstyle \pm 0.2740$ & $0.4444 \scriptstyle \pm 0.3143$ & \boldmath{$1.0000 \scriptstyle \pm 0.0000$} \\
\midrule
\multirow{2}{*}{Others$^{\ast}$} & Accuracy & $0.5000 \scriptstyle \pm 0.0000$ & $0.5185 \scriptstyle \pm 0.0262$ & $0.4444 \scriptstyle \pm 0.0454$ & $0.4815 \scriptstyle \pm 0.0262$ & \boldmath{$0.8148 \scriptstyle \pm 0.1048$} \\
 & F1 & $0.2222 \scriptstyle \pm 0.3143$ & $0.5136 \scriptstyle \pm 0.2348$ & $0.3961 \scriptstyle \pm 0.2863$ & $0.4274 \scriptstyle \pm 0.3029$ & \boldmath{$0.8185 \scriptstyle \pm 0.1075$} \\
\bottomrule
\multicolumn{7}{l}{\footnotesize Others$^{\ast}$: Endocardial cell; Lymphocyte; Macrophage; Pericyte; Schwann cell} \\
\end{tabular}%
}
\end{center}
\vspace{-0.05in}
\end{table}

\newpage

\section{Supplementary Figures}
\setcounter{figure}{0}
\makeatletter 
\renewcommand{\thefigure}{S\@arabic\c@figure}
\makeatother

\begin{figure}[h!]
\vspace{-0.0in}
\centering
\captionsetup{skip=2.5pt} 
\includegraphics[width=0.85\textwidth]{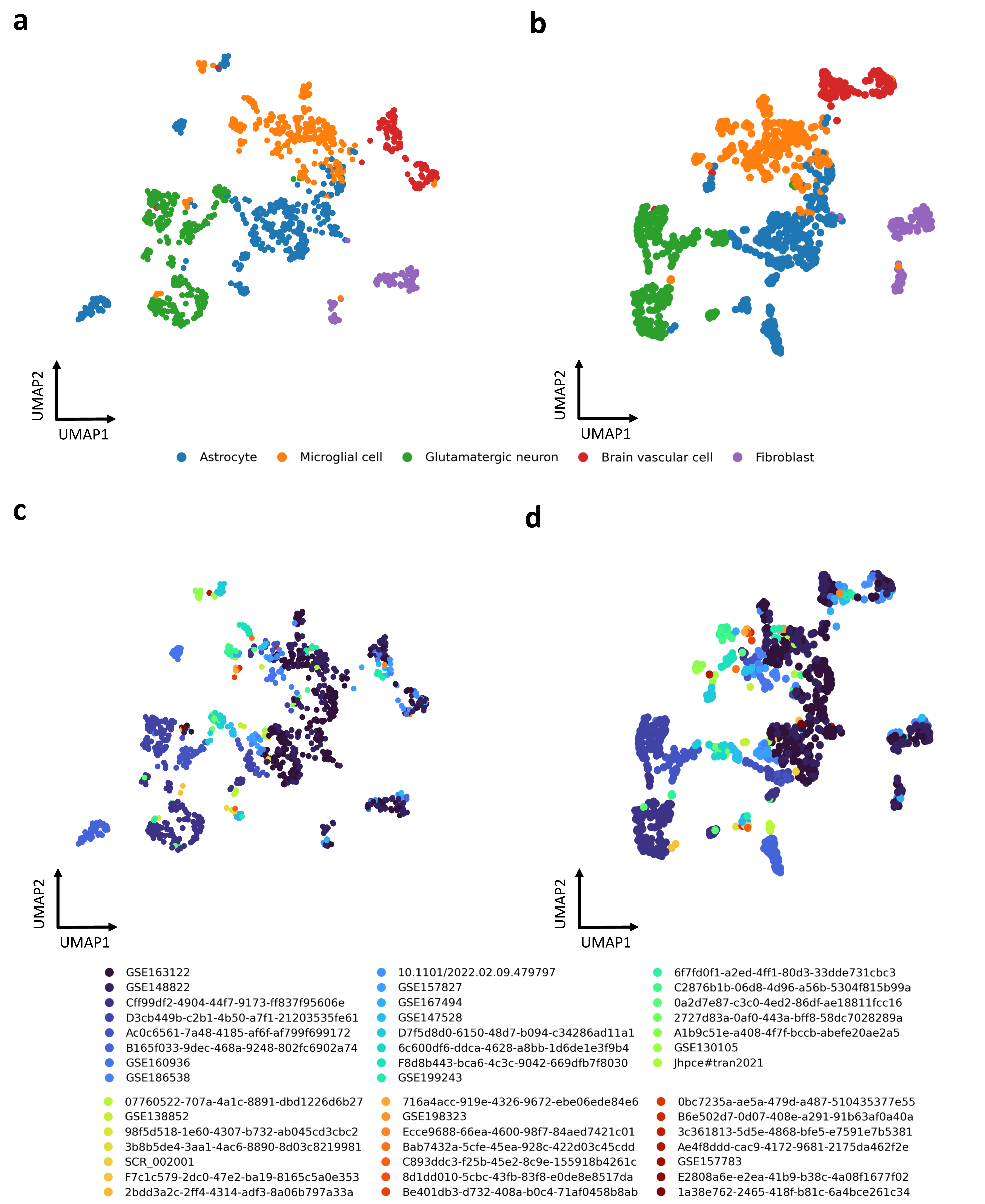}
\caption{\textbf{UMAP embeddings of AD cells before and after ComBat correction.}
\textbf{(a)} Cell-type distribution (pre-ComBat); \textbf{(b)} cell-type distribution (post-ComBat);
\textbf{(c)} data-source distribution (pre-ComBat); \textbf{(d)} data-source distribution (post-ComBat).}
\label{fig:combat-results}
\vspace{-0.25in}
\end{figure}

\newpage
\begin{figure}[h!]
\vspace{-0.0in}
\centering
\captionsetup{skip=2.5pt} 
\includegraphics[width=0.8\textwidth]{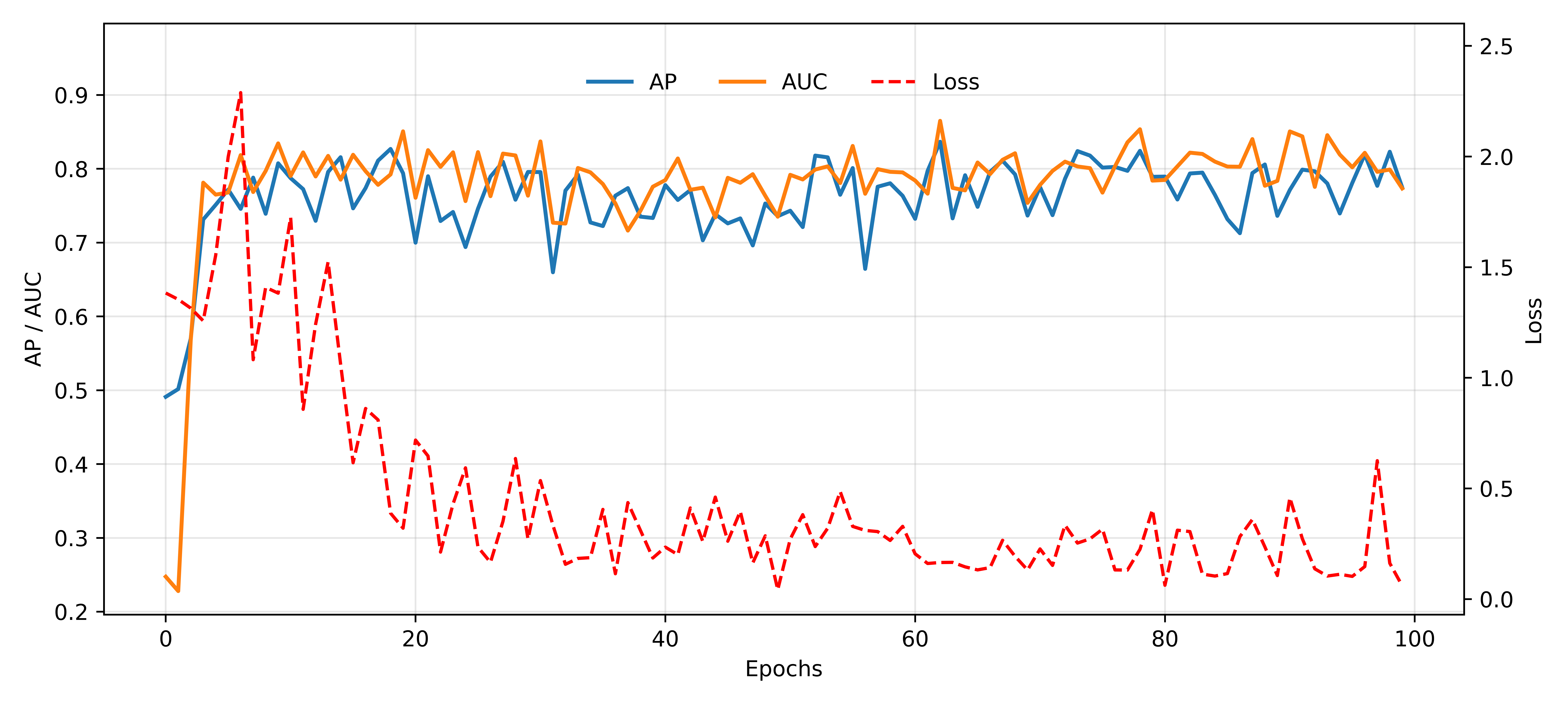}
\caption{\textbf{Experimental results of CellTOSG-FM pretraining}
}
\label{fig:celltosg-fm-performances}
\vspace{-0.25in}
\end{figure}

\newpage
\begin{figure}[h!]
\vspace{-0.0in}
\centering
\captionsetup{skip=2.5pt} 
\includegraphics[width=0.85\textwidth]{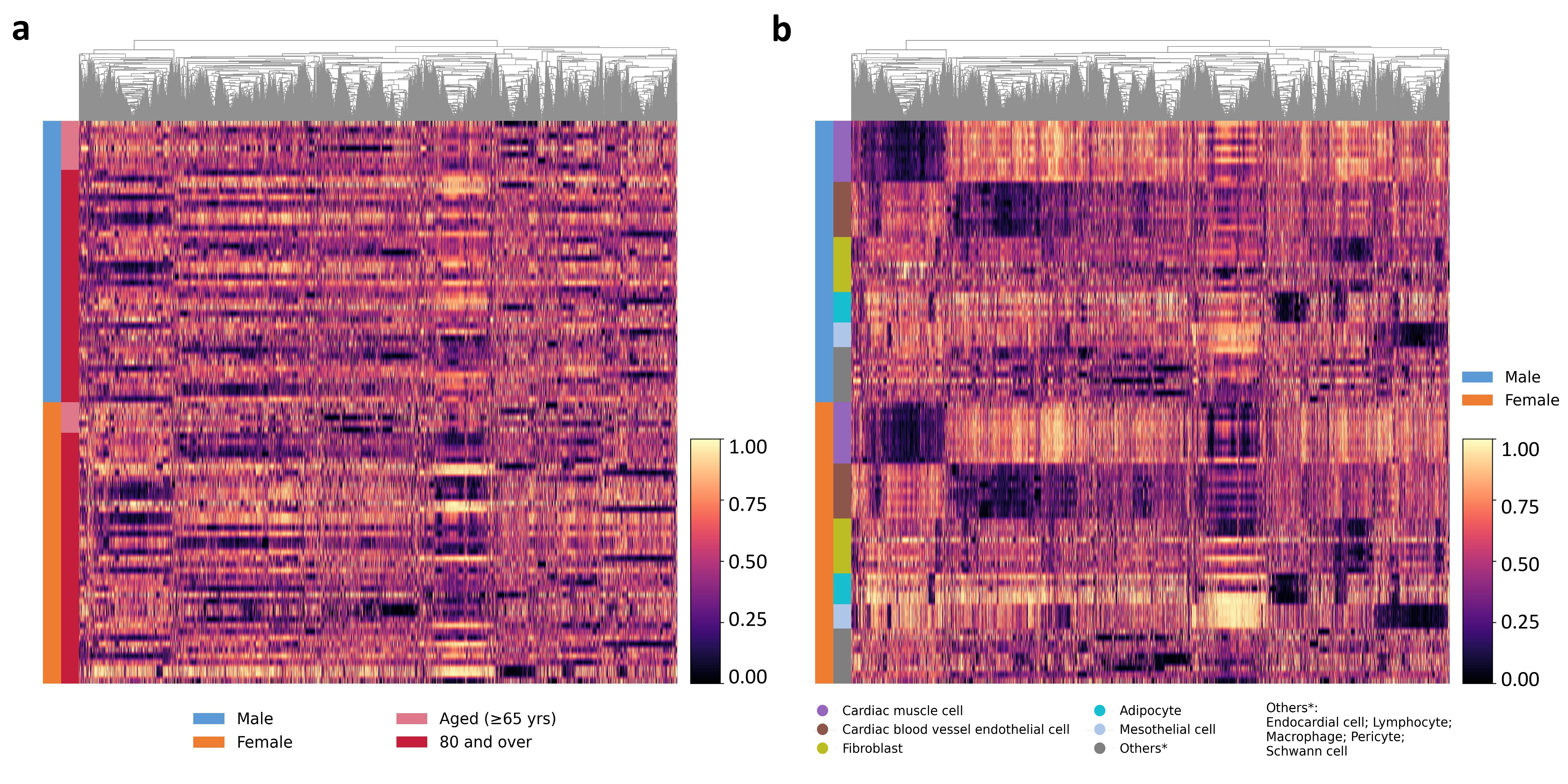}
\caption{\textbf{Cell embeddings learned by CellTOSG-FM for sex classification tasks on AF dataset.}
\textbf{(a)} Cell sex classification task on AF dataset stratified by age groups.
\textbf{(b)} Cell sex classification task on AF dataset stratified by cell types.
}
\label{fig:cell-embed}
\vspace{-0.25in}
\end{figure}

\end{document}